\documentclass[sigconf]{acmart}
\usepackage{latexsym}
\usepackage{url}

\usepackage{xspace}
\usepackage{graphicx}
\usepackage{multirow}
\usepackage{makecell}
\usepackage{lscape}
\usepackage{bm}
\usepackage{algorithm}
\usepackage{algpseudocode}
\usepackage{framed}
\usepackage{color}
\usepackage{xcolor}
\usepackage{booktabs,colortbl}
\usepackage{bbm}
\usepackage{subfigure}
\usepackage{multirow}

\usepackage{amssymb}
\usepackage{amsmath}
\usepackage{multirow}
\usepackage{makecell}
\usepackage{tabularx}
\usepackage{booktabs}
\usepackage{setspace}
\usepackage{utfsym}
\usepackage{pifont}
\usepackage{arydshln}

\definecolor{lightgrayv}{HTML}{F4F3F8} 
\definecolor{lightbluev}{HTML}{F5F9FD} 
\definecolor{grayv}{HTML}{707070}

\newcommand{\eg}{\emph{e.g.,}\xspace}
\newcommand{\ie}{\emph{i.e.,}\xspace}

\newcommand{\baby}{\textsc{CameRed}\xspace}
\AtBeginDocument{%
  \providecommand\BibTeX{{%
    \normalfont B\kern-0.5em{\scshape i\kern-0.25em b}\kern-0.8em\TeX}}}

\copyrightyear{2025}
\acmYear{2025}
\setcopyright{acmlicensed}
\acmConference[SIGIR '25] {Proceedings of the 48th International ACM SIGIR Conference on Research and Development in Information Retrieval}{ July 13--18, 2025}{Padua, Italy.}
\acmBooktitle{Proceedings of the 48th International ACM SIGIR Conference on Research and Development in Information Retrieval (SIGIR '25), July 13--18, 2025, Padua, Italy}
\acmISBN{979-8-4007-1592-1/25/07}
\acmDOI{10.1145/XXXXXX.XXXXXX}

\begin{document}

\title{\textit{Collaboration and Controversy Among Experts}: \\ Rumor Early Detection by Tuning a Comment Generator}

\author{Bing Wang}
\orcid{0000-0002-1304-3718}
\affiliation{
  \institution{College of Computer Science and Technology, Jilin University}
  \city{Changchun}
  \state{Jilin}
  \country{China}}
\email{wangbing1416@gmail.com}

\author{Bingrui Zhao}
\affiliation{%
  \institution{College of Computer Science and Technology, Jilin University}
  \city{Changchun}
  \state{Jilin}
  \country{China}}
\email{zhaobr02@163.com}

\author{Ximing Li}
\thanks{* Ximing Li are corresponding authors. All authors are affiliated with Key Laboratory of Symbolic Computation and Knowledge Engineering of the Ministry
 of Education, Jilin University.}
\authornotemark[1]
\orcid{0000-0001-8190-5087}
\affiliation{
  \institution{College of Computer Science and Technology, Jilin University}
  \city{Changchun}
  \state{Jilin}
  \country{China}}
\email{liximing86@gmail.com}

\author{Changchun Li}
\orcid{0000-0002-8001-2655}
\affiliation{
  \institution{College of Computer Science and Technology, Jilin University}
  \city{Changchun}
  \state{Jilin}
  \country{China}}
\email{changchunli93@gmail.com}

\author{Wanfu Gao}
\orcid{0000-0003-2738-596X}
\affiliation{
  \institution{College of Computer Science and Technology, Jilin University}
  \city{Changchun}
  \state{Jilin}
  \country{China}}
\email{gaowf@jlu.edu.cn}

\author{Shengsheng Wang}
\orcid{0000-0002-8503-8061}
\affiliation{
  \institution{College of Computer Science and Technology, Jilin University}
  \city{Changchun}
  \state{Jilin}
  \country{China}}
\email{wss@jlu.edu.cn}

\renewcommand{\shortauthors}{Bing Wang et al.}

\begin{abstract}
  Over the past decade, social media platforms have been key in spreading rumors, leading to significant negative impacts. To counter this, the community has developed various \textbf{R}umor \textbf{D}etection (\textbf{RD}) algorithms to automatically identify them using user comments as evidence. However, these RD methods often fail in the early stages of rumor propagation when only limited user comments are available, leading the community to focus on a more challenging topic named \textbf{R}umor \textbf{E}arly \textbf{D}etection (\textbf{RED}).
Typically, existing RED methods learn from limited semantics in early comments. However, our preliminary experiment reveals that the RED models always perform best when the number of training and test comments is consistently extensive. This inspires us to address the RED issue by generating more human-like comments to support this hypothesis. To implement this idea, we tune a comment generator by simulating expert collaboration and controversy and propose a new RED framework named \baby. Specifically, we integrate a mixture-of-expert structure into a generative language model and present a novel routing network for expert collaboration. Additionally, we synthesize a knowledgeable dataset and design an adversarial learning strategy to align the style of generated comments with real-world comments. We further integrate generated and original comments with a mutual controversy fusion module. Experimental results show that \baby outperforms state-of-the-art RED baseline models and generation methods, demonstrating its effectiveness.
\end{abstract}




\maketitle

\section{Introduction}

Currently, the uncontrolled proliferation of rumors exist on various social media platforms, \eg Twitter and Weibo. People make comments on them and spread them, posing significant threats to cybersecurity and the safety of citizens' property \citep{vosoughi2018spread}. For example, recent rumors claim that "\textit{a certain cryptocurrency will surge dramatically in the next few months}," leading to excessive speculation among cryptocurrency investors.\footnote{\url{https://www.cnbc.com/2022/01/11/crypto-scams-are-the-top-threat-to-investors-by-far-say-regulators.html}.} 
The ones who believed this rumor, have suffered substantial losses.
To tackle this issue and block the spread of rumors, researchers have developed automatic \textbf{R}umor \textbf{D}etection (\textbf{RD}) algorithms, which aim to promptly identify and locate rumors on social media platforms, thereby facilitating the implementation of effective countermeasures to mitigate their impact \citep{zuo2022continually,chen2022cross,wang2024why}.

\begin{figure}[t]
  \centering
  \includegraphics[scale=0.40]{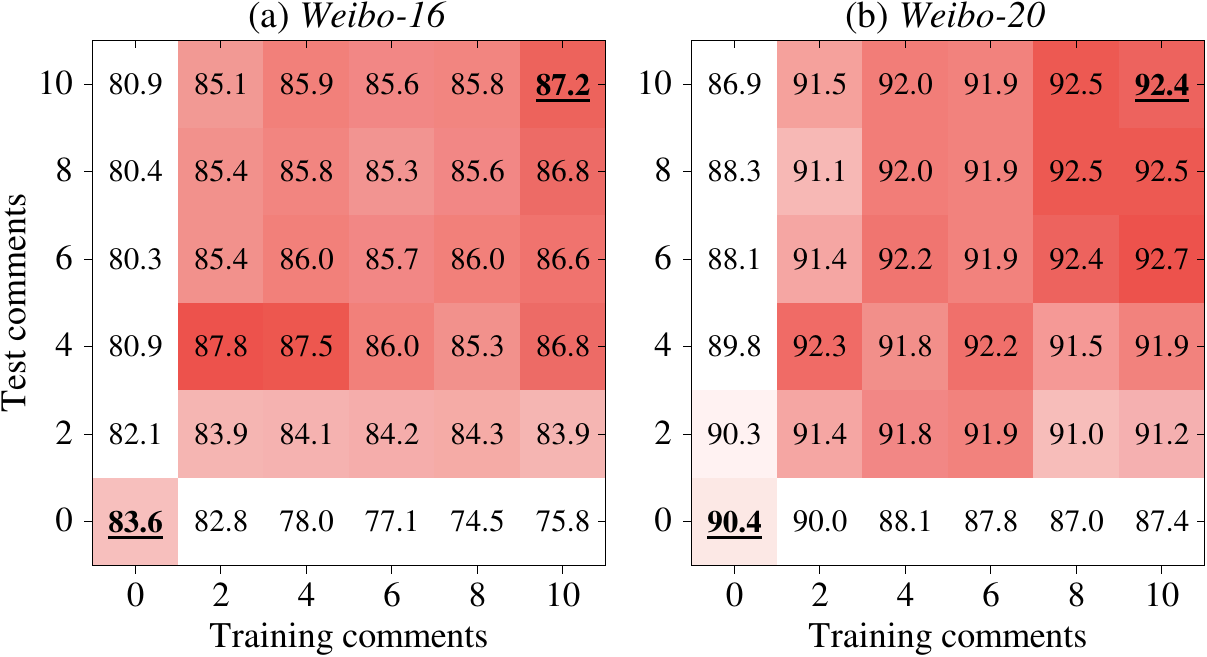}
  \caption{We report Macro F1 metrics when varying the number of training and test comments in two RD datasets.}
  \label{introduction}
\end{figure}

Generally, cutting-edge RD arts collect posts and corresponding user engagements, \eg user comments, and then encode them into a hidden semantic space to capture the underlying relationships between the semantics and their veracity labels, \eg real and fake \citep{zuo2022continually,sun2023hg,lin2023zero}. For example, some works organize user comments into a tree structure, and design various graph neural networks \citep{kipf2017semi} to map the comments into discriminative features \citep{ma2018rumor,sun2023hg,lin2023zero}.

Typically, these RD models always assume that user comments are sufficient to support the detection. However, in the early stage of the post propagation, user engagement is limited, resulting in few or even no comments available. Therefore, detecting rumors during their early stages presents a more challenging problem, named \textbf{R}umor \textbf{E}arly \textbf{D}etection (\textbf{RED}) \citep{wang2021early,dong2022improving,zhou2022mamn}. To tackle this challenge, current RED methods employ neural processing \citep{zeng2022early} and temporal forecasting \citep{song2021ced,zeng2022early,li2022adadebunk} to learn limited semantics from early comments.

Despite the effectiveness of existing RED methods \citep{zeng2022early,li2022adadebunk,nan2024exploiting,zhang2024mitigating}, the limitedly available comments consistently hampers the peformance of detection models. We present a preliminary experiment with a SOTA RED method to observe the issue as depicted in Fig.~\ref{introduction}. On one hand, when validated on practical applications, the scarcity of test comments on unknown posts prevents the model from extracting meaningful features. On the other hand, the limited number of comments in the training data also leads to insufficient model training. Upon the observations, we assume that the model performs best when the number of training and test comments is consistently extensive.

Motivated by our preliminary experiment, a natural idea to address the RED issue is to generate human-like comments to keep the comments in training and test phases consistently extensive. Accordingly, to implement this idea, we draw inspiration from the \textbf{C}ollaboration and \textbf{C}ontroversies \textbf{AM}ong human \textbf{E}xperts, and propose a new \textbf{RED} framework, named \textbf{\baby}, which tunes a comment generator to simulate the collaboration of multiple experts and integrates generated comments by learning their controversy viewpoints.
Specifically, the basic ideas of \baby are two-fold. First, we incorporate a Mixture-of-Experts (MoE) structure \citep{shen2019mixture,dou2024loramoe} into a pre-trained language model, \eg T5 \citep{chung2022scaling} and Llama \citep{touvron2023llama}. Each expert model is tuned on different user comments and synthetic knowledgeable corpora. To align the style of the generated comments with that of real-world comments, we propose an adversarial strategy to optimize the expert models. On the other hand, we integrate the original tweet comments with the generated ones through a mutual controversy fusion module, which groups comments by their stance toward the tweet and combines the semantic features of the comments within each group.

We compare \baby with five baseline RED models across four prevalent RED datasets \textit{Twitter15} \citep{ma2017detect}, \textit{Twitter16} \citep{ma2017detect}, \textit{Weibo16} \citep{ma2016detecting} and \textit{Weibo20} \citep{zhang2021mining}. The experimental results demonstrate that \baby consistently and significantly improves the baseline model performance. Additionally, we compare our generated comments with those from two SOTA RED methods, DELL \citep{wan2024dell} and GenFEND \citep{nan2024let}, and find that \baby outperform them in terms of diversity, style alignment, and improvement of detection model performance. The source code and data are released in \textit{\url{https://github.com/wangbing1416/CAMERED}}.

The primary contributions of this work can be summarized into the following three-folds:
\begin{itemize}
    \item Our preliminary experiment shows that in RED, the model performs best when the number of training and testing comments is consistently extensive. Therefore, we propose to address the RED issue by generating human-like comments.
    \item We propose a new RED framework \baby that tunes a comment generator and integrates the generated comments.
    \item Extensive experiments on four prevalent RED datasets and two RED scenarios are conducted to prove the effectiveness of the \baby method.
\end{itemize}

\section{Related Works}

In this section, we briefly review the relevant literature about RED and the pre-trained text generators.

\subsection{Rumor Early Detection}

Generally, cutting-edge RD models learn the underlying relationship between veracity labels and various features of posts, \eg semantic information \citep{chen2022cross,wang2024harmfully,wang2024escaping}, intent features \citep{wang2023understanding,wang2024why}, and emotional signals \citep{zhang2021mining}, with advanced deep models. Meanwhile, the social context is also a crucial aspect of RD, which induces a series of graph-based methods to learn key features from information propagation structures \citep{sun2023hg,lin2023zero}.
Despite the success of the RD methods, they still struggle to handle the scenario of detecting early rumors before their rapid spreads, therefore raising a challenging RED task. 
Typically, current RED arts are roughly classified into the following three categories. \textbf{Adaptation methods} \citep{wang2018eann,choi2021using,zhang2024evolving}, capture event-invariant features to adapt the model trained across old events to new events. \textbf{Temporal methods} \citep{song2021ced,zuo2022continually,li2022adadebunk}, learn the varying trend of prediction confidence as the comments increase, and select the point with the highest confidence as the final prediction. \textbf{Weakly-supervised methods} \citep{wang2021multimodal,lin2022detect,zeng2022early}, regard RED as a weakly-supervised learning problem on low-resource new events, and resort to prevalent learning methods, \eg neural process.

With the recent development of the LLM technique, some recent RD works enhance the detection model by generating various aspects, \eg text descriptions, commonsense rationales, and user reactions \citep{he2023reinforcement,hu2024bad,wan2024dell,yue2024evidence,nan2024let}. Typically, they directly load a general-purpose LLM for generation, which consistently struggles to produce real-world comments.

\subsection{Pre-trained Text Generators}

Generally, current text generators always resort to large-scale pre-trained language models, \eg T5 \citep{chung2022scaling} and Llama \citep{touvron2023llama}, which generate responses by an auto-regressive regime. Our work tunes a generator aiming to produce diverse, knowledgeable, and human-like comments, and we review the related methods in this section.

\noindent
\textbf{Diverse generation.}
Current works primarily generate diverse responses from the encoder and decoder perspectives. On the encoder side, recent works propose parameter isolation methods, \eg mixtures of encoders \citep{he2018sequence,shen2019mixture} or paragraph selectors \citep{cho2019mixture}, to produce multiple outputs. Moreover, limited by the fixed model architecture of LLMs, cutting-edge works design various decoding algorithms, \eg beam search \citep{vijayakumar2016diverse}, nucleus sampling \citep{holtzman2020the}, and contrastive decoding \citep{li2023contrastive}, to sample the logits of LLM outputs, encouraging them to generate fluent and diverse responses.

\noindent
\textbf{Knowledge injection.} 
To enable LLMs to encapsulate extensive factual information, a variety of methods have been developed to inject external knowledge into LLMs. These methods can be broadly categorized into explicit and implicit approaches. Explicit methods involve retrieving relevant claims directly from the Internet and inputting them to LLMs for appropriate summarization, \ie retrieval-augmented generation techniques \citep{gao2023retrieval,wang2024m,zeng2024justilm}. On the other hand, implicit methods construct large knowledge corpora and design new training objectives to post-train the LLM \citep{zhou2019gear,zhang2023plug}.

\noindent
\textbf{Style alignment.}
In the visual generation community, recent work aligns styles by extracting latent features, \eg pose and identity, from target images and using them as priors for styled image generation \citep{karras2019a,yang2023styleganex}. A related concept in LLMs involves recent human preference alignment approaches, which guide LLMs to produce outputs that better match human preferences \citep{christiano2017deep,ouyang2022training}. Inspired by adversarial preference alignment \citep{cheng2023adversarial}, our work proposes to employ adversarial learning to train the comment generator.

\section{Our Proposed Method}

In this section, we present the problem formulation of RED and our proposed method \baby in detail.

\vspace{3pt} \noindent
\textbf{Problem formulation.}
Typically, let $\mathcal{D} = \{\mathbf{x}_i, \mathcal{C}_i, y_i\}_{i=1}^N$ denote an RED dataset, each sample consists of a post $\mathbf{x}_i$, its associated user comments $\mathcal{C}_i = \{\mathbf{c}_{ij}\}_{j=1}^M$, and a veracity label $y_i \in \{0, 1\}$, \eg real and fake. Here, $N$ denotes the size of the training dataset, and $M$ represents the number of comments per post. 
Generally, the goal of RED is to train a detector to predict veracity labels $y_i^\prime$ for the previously unseen post $\{\mathbf{x}_i^\prime, \mathcal{C}_i^\prime\}$, where $\mathcal{C}_i^\prime =\{\mathbf{c}_{ij}^\prime\}_{j=1}^{M^\prime}$. In the RED scenario, the number of comments $M^\prime$ in the test set is consistently limited, even $M^\prime = 0$, which leads to sub-optimal performance (as proven in Fig.~\ref{introduction}). 
To address this issue, our work tunes a comment generator to produce $K$ additional comments for the training set and $K^\prime$ comments for the test set to keep $M + K = M^\prime + K^\prime$.

\subsection{Overview of \baby}

\begin{figure*}[t]
  \centering
  \includegraphics[scale=0.64]{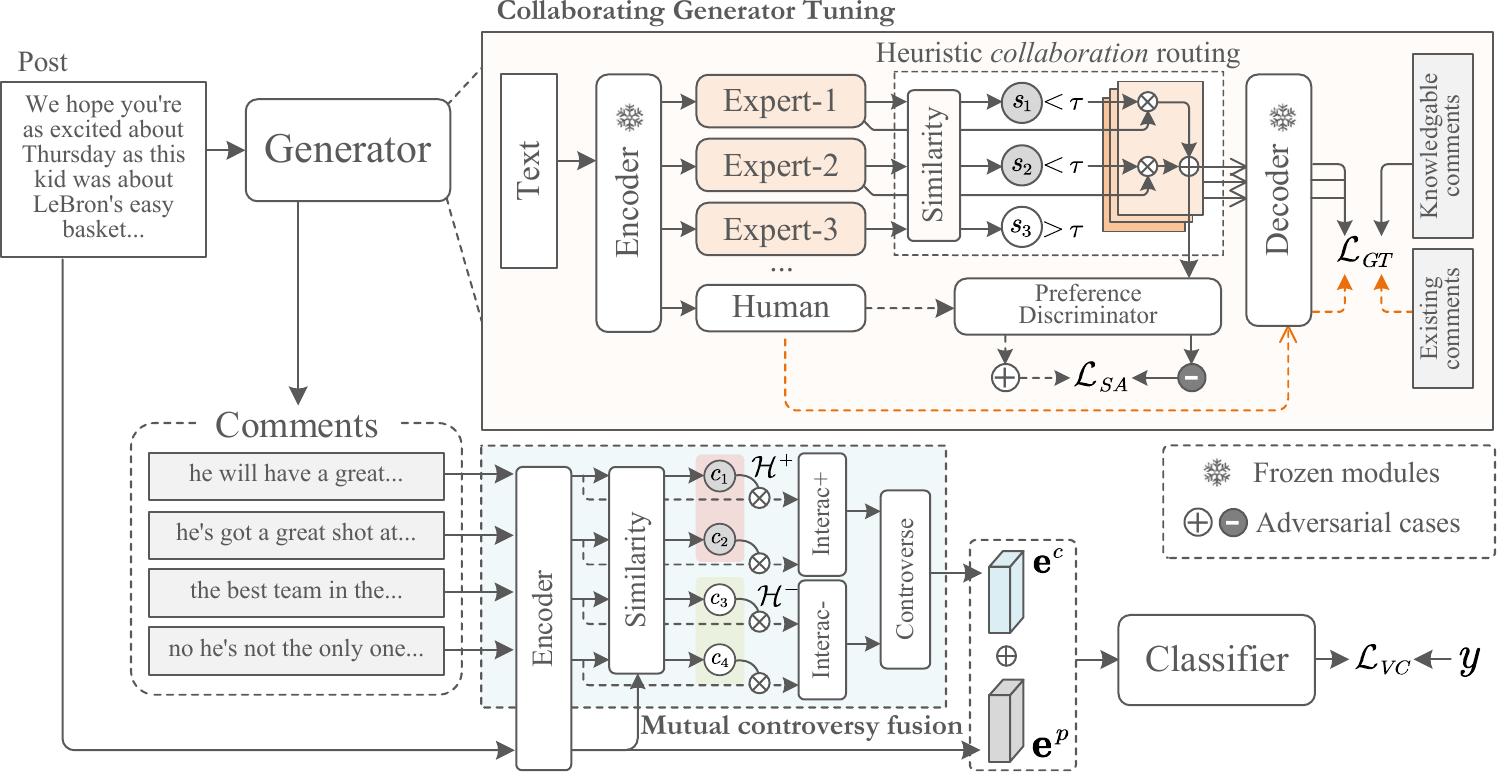}
  \caption{Overall framework of \baby. We first tune a generator by integrating multiple experts. Then, the generated comments, along with the original comments, are feed into a mutual controversy fusion module for prediction.}
  \label{framework}
\end{figure*}

We draw inspiration from the empirical observation that the RED model achieves optimal performance when the number of comments during training and testing is consistent and extensive. However, in RED, the number of comments available during testing is always insufficient to support these observations. This naturally leads us to the idea of tuning a comment generator to fill this gap. 
Upon this idea, inspired by the processes of \textit{collaboration} and \textit{controversy} among human experts, we design a comment generator based on the mixture-of-experts structure, which produces diverse, knowledgeable, and human-like comments through expert collaboration. Additionally, we develop a new comment integration model that simulates the controversies among experts to derive the final judgment. Accordingly, we propose a new RED framework \baby, which consists of three basic modules: \textbf{collaborating generator tuning}, \textbf{mutual controversy fusion}, and \textbf{veracity classifier}. The overall framework is depicted in Fig.~\ref{framework}. In the following, we briefly describe these modules.

\vspace{2pt} \noindent
\textbf{Collaborating generator tuning.}
This module trains the comment generator $\mathcal{G}_{\boldsymbol{\pi}}(\cdot)$ across $\{\mathbf{x}_i, \mathcal{C}_i\}_{i=1}^N \in \mathcal{D}$ based on the pre-trained weights of two language models: \textit{Flan-T5} ($\sim$220M) \citep{chung2022scaling} and \textit{Llama} ($\sim$7B) \citep{touvron2023llama}. 
To ensure that the generator produces diverse and knowledgeable comments, we integrate a mixture-of-experts module \citep{shen2019mixture,dou2024loramoe} into the pre-trained model, and train each expert model with a synthetic dataset $\mathcal{D}_\kappa = \{\mathbf{x}_i^\kappa, \mathbf{c}_i^\kappa \}_{i=1}^{|\mathcal{D}_\kappa|}$ enriched with entity knowledge.
To simulate collaboration among experts, we ground the hypothesis that \textit{experts with similar viewpoints can collaborate more effectively} \citep{smith2021great,wang2024explainable} and design a collaborative routing strategy for the mixture-of-experts model.
Additionally, we propose an adversarial optimization strategy \citep{cheng2023adversarial} to align the style of generated comments with human preferences.

\vspace{2pt} \noindent
\textbf{Mutual controversy fusion.}
Given a post $\mathbf{x}_i$, the tuned generator produces $K$ comments $\{\mathbf{c}_{ij}^\gamma\}_{j=1}^K = \mathcal{G}_{\boldsymbol{\pi}}(\mathbf{x}_i)$. These generated comments, along with the previous comments $\{\mathbf{c}_{ij}\}_{j=1}^M$, are then input into a \textit{mutual controversy fusion} module $\mathcal{F}_{\boldsymbol{\theta}}(\cdot)$ to generate a comment feature $\mathbf{e}_i^c$. 
Specifically, we divide the comment set $\{\mathbf{c}_{ij}^\gamma\}_{j=1}^K \cup \{\mathbf{c}_{ij}\}_{j=1}^M$ into two subsets based on their stances and analyze the differences between the two subsets to simulate the controversies among experts.

\vspace{2pt} \noindent
\textbf{Veracity classifier.}
Accordingly, we extract the semantic feature $\mathbf{e}_i^p$ of the post $\mathbf{x}_i$ with BERT \citep{devlin2019bert} and concatenate it with $\mathbf{e}_i^c$ as input to a \textit{veracity classifier} to predict veracity labels. The classification objective across the dataset $\mathcal{D}$ is formulated as follows:
\begin{equation}
    \label{eq1}
    \mathcal{L}_{VC} = \frac{1}{N} \sum \nolimits _{i=1}^N \ell_{CE} \Big( \big[ \mathbf{e}_i^p; \mathbf{e}_i^c \big] \mathbf{W}_C, y_i \Big),
\end{equation}
where $\mathbf{W}_C$ is a veracity classifier, and $[\cdot\ ;\cdot]$ and $\ell_{CE}(\cdot\ , \cdot)$ represent a concatenation operation and a cross-entropy loss function, respectively. In the following sections, we describe \textit{collaborating generator tuning} and \textit{mutual controversy fusion} modules in more detail.

\subsection{Collaborating Generator Tuning}

Generally, the collaborating generator tuning module involves tuning a comment generator $\mathcal{G}_{\boldsymbol{\pi}}(\cdot)$ across $\mathcal{D}$, which enables the generated comments satisfy diverse, knowledgeable, and human-like advantages. 
To achieve this goal, we implement the three strategies: \textit{multiple experts structure}, \textit{knowledgeable data synthesis}, and \textit{adversarial style alignment}.

\vspace{2pt} \noindent
\textbf{Multiple experts structure.}
Typically, a pre-trained language model can be decoupled as an encoder (or an embedding layer in Llama) formulated as  $\mathcal{G}_{\boldsymbol{\pi}_e}(\cdot)$ and a decoder $\mathcal{G}_{\boldsymbol{\pi}_d}(\cdot)$. 
Given a social media post $\mathbf{x}_i$, we input it into the encoder to obtain the hidden embeddings $\mathbf{H}_i = \mathcal{G}_{\boldsymbol{\pi}_e}(\mathbf{x}_i)$. To enable a single model to generate a diverse array of comments, we freeze the pre-trained language model and efficiently tune a mixture-of-experts structure \citep{shen2019mixture,dou2024loramoe}. 
Specifically, we design $L$ expert models $\{\mathcal{E}_{\boldsymbol{\phi}_l}(\cdot)\}_{l=1}^L$, and the hidden embedding $\mathbf{H}_i$ from the encoder is then fed into these experts to obtain expert embeddings $\mathbf{h}_{il} = \mathcal{E}_{\boldsymbol{\phi}_l}(\mathbf{H}_i)$, where $l \in \{1, \cdots, L\}$. 

Afterward, we develop a heuristic collaboration routing strategy to simulate the collaboration among experts. We take inspiration from the fact that \textit{groups of experts holding similar viewpoints can cooperate more efficiently} \citep{smith2021great,wang2024explainable}. To implement this, we organize similar experts into one group and randomly combine experts within this group. Formally, based on the expert embeddings, we calculate pairwise similarities between experts and construct a similarity matrix as follows:
\begin{equation}
    \label{eq2}
    \mathbf{A}_i = \{a_{ilm}\}_{l, m \in \{1, \cdots, L\}}, \quad a_{ilm} = \frac{\mathbf{h}_{il} \cdot \mathbf{h}_{im}}{\|\mathbf{h}_{il}\| \times \|\mathbf{h}_{im}\|}.
\end{equation}

Naturally, $\mathbf{A}_i$ can be interpreted as the adjacency matrix of a weighted undirected graph, where $\mathbf{h}_{il}$ and $a_{ilm}$ represent a node and an edge, respectively. Then, we prune the graph by removing edges with low similarity scores, and set their weights to zero, as:
\begin{equation}
    \label{eq3}
    a_{ilm} \gets 0, \quad \text{if}\ a_{ilm} < \epsilon, \nonumber
\end{equation}
where $\epsilon$ is a manual threshold. After pruning, the disconnected sub-graphs in $\mathbf{A}_i$ are mutually independent, and their expert embeddings can be organized into $T$ non-overlapping groups $\{\mathcal{H}_i^t\}_{t=1}^T$. In each group $\mathcal{H}_i^t$, we randomly select $s \in \{1, \cdots, | \mathcal{H}_i^t | \}$ experts, and average their embeddings to create an internal embedding $\mathbf{o}_i$.
Given $\mathbf{o}_i$, the decoder operates in an auto-regressive regime to generate a comment $\mathbf{c}_i^\gamma = \mathcal{G}_{\boldsymbol{\pi}_d}(\mathbf{o}_i)$. We can repeat this random routing $K$ times to generate $K$ different comments $\{\mathbf{c}_{ij}^\gamma\}_{j=1}^K$. In summary, we can have a total of $\sum \nolimits _{t=1}^T \big( 2^{| \mathcal{H}_i^t |} - 1 \big)$ different combinations of experts, which not only diversify the model’s output but also achieve the collaboration among experts. 

\vspace{3pt} \noindent
\textbf{Knowledgeable data synthesis.}
To ensure that each group of experts specializes in different expertise, we synthesize a knowledgeable dataset $\mathcal{D}_\kappa = \{\mathbf{x}_i, \mathbf{c}_i^\kappa \}_{i=1}^{|\mathcal{D}_\kappa|}$ to train each group of experts, respectively.
Specifically, given $\mathbf{x}_i \in \mathcal{D}$, we extract the entities and their descriptions with an off-the-shelf knowledge graph toolkit \textit{SpaCy}, and then summarize these descriptions into text segments that are approximately the same length as $\mathbf{c}_i$, which are denoted as $\mathbf{c}_i^\kappa$.
Based on the dataset $\mathcal{D} \cup \mathcal{D}_\kappa$, the tuning objective of the generator can be formulated as:
\begin{align}
    \label{eq4}
    \small
    \mathop{\boldsymbol{\min}} \limits _{\boldsymbol{\phi}_{1:L}} \mathcal{L}_{GT} = \frac{1}{NM} \sum \nolimits _{i=1}^N \sum \nolimits _{j=1}^M \frac{1}{|\mathbf{c}_{ij}|} \sum \nolimits _{k=1}^{|\mathbf{c}_{ij}|}
    \ell_{CE} \Big( \mathcal{G}_{\boldsymbol{\pi}} \big( \mathbf{x}_i, \mathbf{c}_{ij<k} \big), & c_{ijk} \Big) \nonumber \\
    + \frac{1}{|\mathcal{D}_\kappa|} \sum \nolimits _{i=1}^{|\mathcal{D}_\kappa|} \frac{1}{|\mathbf{c}_i^\kappa|} \sum \nolimits _{k=1}^{|\mathbf{c}_i^\kappa|}
    \ell_{CE} \Big( \mathcal{G}_{\boldsymbol{\pi}} \big( \mathbf{x}_i, \mathbf{c}_{i<k}^\kappa \big), c_{ik}^\kappa & \Big).
\end{align}

\noindent
\textbf{Adversarial style alignment.}
In our experiments, we observe that comments with inconsistent language styles can lead to a decline in detection performance. Accordingly, to align the style of the comments generated by $\mathcal{G}_{\boldsymbol{\pi}}(\cdot)$ with comments from the target dataset $\mathcal{D}$, we draw inspiration from adversarial preference optimization \citep{cheng2023adversarial} to design an adversarial style alignment approach. 

Specifically, in addition to $L$ expert models, we introduce an additional expert model $\mathcal{E}_{\boldsymbol{\phi}_H}(\cdot)$, which is trained only across the original dataset $\{\mathbf{x}_i, \mathcal{C}_i\} \in \mathcal{D}$, to simulate the language style of humans in $\mathcal{D}$ unaffected by the knowledgeable dataset $\mathcal{D}_\kappa$. Its training objective is formulated as follows:
\begin{equation}
    \label{eq5}
    \small
    \mathop{\boldsymbol{\min}} \limits _{\boldsymbol{\phi}_H} \mathcal{L}_{AM} = \frac{1}{NM} \sum \nolimits _{i=1}^N \sum \nolimits _{j=1}^M \frac{1}{|\mathbf{c}_{ij}|} \sum \nolimits _{k=1}^{|\mathbf{c}_{ij}|} \ell_{CE} \Big( \mathcal{G}_{\boldsymbol{\pi}} \big( \mathbf{x}_i, \mathbf{c}_{ij<k} \big), c_{ijk} \Big).
\end{equation}
Based on it, we expect the style feature of expert embedding $\mathbf{o}_{i}$ is close to that of $\mathbf{h}_i^H = \mathcal{E}_{\boldsymbol{\phi}_H}(\mathbf{H}_i)$.
To achieve this, we train a style discriminator $\mathbf{W}_S$ to predict the style labels as 0/1, where 1 indicates that the style of the input embedding is closer to comments written by humans in $\mathcal{D}$, and vice versa. Meanwhile, the expert models $\mathcal{E}_{\boldsymbol{\phi}_{1:L}}(\cdot)$ aim to fool the style discriminator to confuse the embeddings with different styles. Accordingly, we propose the adversarial objective denoted as the following MIN-MAX formula:
\begin{equation}
    \label{eq6}
    \small
    \mathop{\boldsymbol{\max}} \limits _{\boldsymbol{\phi}_{1:L}} \mathop{\boldsymbol{\min}} \limits _{\mathbf{W}_S} \mathcal{L}_{SA} = \frac{1}{N + |\mathcal{D}_\kappa|}
    \sum \nolimits _{i=1}^{N + |\mathcal{D}_\kappa|} \ell_{CE} \big( \mathbf{o}_i \mathbf{W}_S, 0 \big) + \ell_{CE} \big( \mathbf{h}_i^H \mathbf{W}_S, 1 \big).
\end{equation}
To efficiently implement the adversarial training, we leverage a gradient reverse layer \citep{ganin2015unsupervised}, which reverses the gradient $\frac{\partial \mathcal{L}_{SA}}{\partial \boldsymbol{\phi_{1:L}}}$ to $- \frac{\partial \mathcal{L}_{ASA}}{\partial \boldsymbol{\phi_{1:L}}}$ during the backpropagation process.

In summary, the overall objective of our proposed collaborating generator tuning module is Eq.~\eqref{eq7}. 
\begin{equation}
    \label{eq7}
    \mathop{\boldsymbol{\max}} \limits _{\boldsymbol{\phi_{1:L}}} \mathop{\boldsymbol{\min}} \limits _{\{\boldsymbol{\phi_{1:L}}, \boldsymbol{\phi_H}, \mathbf{W}_S\}}
    \mathcal{L}_{CGT} = \mathcal{L}_{GT} + \alpha \mathcal{L}_{AM} + \beta \mathcal{L}_{SA},
\end{equation}
where $\alpha$ and $\beta$ are manual trade-off parameters to balance multiple objectives.

\subsection{Mutual Controversy Fusion}

Given generated comments $\{\mathbf{c}_{ij}^\gamma\}_{j=1}^K$ and original comments $\{\mathbf{c}_{ij}\}_{j=1}^M$, we generate one comment feature $\mathbf{e}_i^c$ utilizing a fusion model parameterized by $\boldsymbol{\theta}$.
Initially, we capture the hidden embeddings $\mathcal{H}_i = \{\mathbf{h}_{ij}^c\}_{j=1}^{M + K}$ from these comments, as well as the embedding $\mathbf{e}_i^p$ of the post $\mathbf{x}_i$, with a pre-trained BERT model \citep{devlin2019bert}.\footnote{Since several studies suggest that average-pooled sentence embeddings outperform ones using \texttt{[CLS]} tokens \citep{li2020on}, we feed a comment piece into BERT and average the token embeddings to form a single hidden embedding.} 
To simulate the controversy, we divide $\mathcal{H}_i$ into two subsets $\mathcal{H}_i^+$ and $\mathcal{H}_i^-$, based on their stances towards the post, \eg support and deny \citep{ma2018detect,yang2022a}. We simply implement this division with semantic similarities between the comment embeddings and the post embedding as follows:
\begin{equation}
    \label{eq8}
    \left\{
    \begin{aligned}
        \ \mathcal{H}_i^+ \gets \mathbf{h}_{ij}^c,& \quad \xi_{ij} > \tau, \\
        \ \mathcal{H}_i^- \gets \mathbf{h}_{ij}^c,& \quad \text{otherwise}.
    \end{aligned}
    \right. 
    \quad \xi_{ij} = 1 - \frac{\mathbf{h}_{ij}^c \cdot \mathbf{e}_i^p}{\|\mathbf{h}_{ij}^c\| \times \|\mathbf{e}_i^p\|},
\end{equation}
where $\tau$ is a manually adjusted parameter that serves to balance two subsets. Subsequently, we respectively extract the subset features of $\mathcal{H}_i^+$ and $\mathcal{H}_i^-$ using self-attention networks, which signify the representatives of the two camps of experts. The features can be formulated as follows:
\begin{equation}
    \label{eq10}
    \mathbf{\widehat h}_{i}^{+} = \mathcal{F}_{\boldsymbol{\theta}_{+}} \left( \big\{ \xi_{ij} \mathbf{h}_{ij}^c \big\}_{\in \mathcal{H}_i^{+}} \right), \quad
    \mathbf{\widehat h}_{i}^{-} = \mathcal{F}_{\boldsymbol{\theta}_{-}} \left( \big\{ \xi_{ij} \mathbf{h}_{ij}^c \big\}_{\in \mathcal{H}_i^{-}} \right),
\end{equation}
where $\boldsymbol{\theta}_{+}$ and $\boldsymbol{\theta}_{-}$ denote the parameters of two self-attention networks. Ultimately, our comment feature $\mathbf{e}_i^c$ is derived from the discrepancy between the two subset features \citep{reimers2019sentence}, and can be computed as follows:
\begin{equation}
    \label{eq11}
    \mathbf{e}_i^c = \left[\mathbf{\widehat h}_{i}^{+}; \mathbf{\widehat h}_{i}^{+} \otimes \mathbf{\widehat h}_{i}^{-}; \mathbf{\widehat h}_{i}^{+} \ominus \mathbf{\widehat h}_{i}^{-}; \mathbf{\widehat h}_{i}^{-}\right] \mathbf{W}_F,
\end{equation}
where $\mathbf{W}_F$ indicates a learnable projection matrix, and $\otimes$ and $\ominus$ represent the feature subtraction and multiplication operations, respectively. In conclusion, the parameter of fusion module $\boldsymbol{\theta} = \{\boldsymbol{\theta}^{+}, \boldsymbol{\theta}^{-}, \mathbf{W}_F\}$.

\section{Experiments}

\begin{table}[t]
\centering
\renewcommand\arraystretch{0.90}
  \caption{The statistics of four RD datasets. \#Num and \#Avg.C. indicate the numbers of samples and comments per sample.}
  \label{dataset}
  \small
  \setlength{\tabcolsep}{5pt}{
  \begin{tabular}{m{1.6cm}<{\centering}m{0.6cm}<{\centering}m{0.8cm}<{\centering}m{0.6cm}<{\centering}m{0.8cm}<{\centering}m{0.6cm}<{\centering}m{0.8cm}<{\centering}}
    \toprule
    \multirow{2}{*}{Dataset} & \multicolumn{2}{c}{Train} & \multicolumn{2}{c}{Validation} & \multicolumn{2}{c}{Test} \\
    \cmidrule(r){2-3} \cmidrule(r){4-5} \cmidrule(r){6-7}
    & \#Num & \#Avg.C. & \#Num & \#Avg.C. & \#Num & \#Avg.C. \\
    \hline
    \textit{Twitter15} \citep{ma2017detect} & 1,133 & 30.62 & 140 & 31.25 & 140 & 30.70 \\
    \textit{Twitter16} \citep{ma2017detect} & 602 & 24.98 & 77 & 33.02 & 77 & 31.45 \\
    \textit{Weibo16} \citep{ma2016detecting} & 2,211 & 511.93 & 738 & 500.13 & 757 & 493.63 \\
    \textit{Weibo20} \citep{zhang2021mining} & 3,816 & 331.74 & 1,272 & 253.95 & 1,274 & 309.63 \\
    \bottomrule
  \end{tabular} }
\end{table}

\begin{table*}[t]
\centering
\renewcommand\arraystretch{0.94}
  \caption{Experimental results under the RED scenario (we fix $M = 16$ and $M^\prime = 2$). The results indicated by * are statistically significant than its baseline model (p-value < 0.05).}
  \label{resultearly}
  \small
  \setlength{\tabcolsep}{5pt}{
  \begin{tabular}{m{1.95cm}m{0.94cm}<{\centering}m{0.94cm}<{\centering}m{0.94cm}<{\centering}m{0.94cm}<{\centering}m{0.94cm}<{\centering}m{0.80cm}<{\centering}m{0.94cm}<{\centering}m{0.94cm}<{\centering}m{0.94cm}<{\centering}m{0.94cm}<{\centering}m{0.94cm}<{\centering}m{0.80cm}<{\centering}}
    \toprule
    \multirow{2}{*}{\quad \quad Model} & \multicolumn{6}{c}{Dataset: \textit{Twitter15} \citep{ma2017detect}} & \multicolumn{6}{c}{Dataset: \textit{Weibo16} \citep{ma2016detecting}} \\
    
    \cmidrule(r){2-7} \cmidrule(r){8-13} 
    & Acc. & F1 & AUC & P. & R. & Avg.~$\boldsymbol{\Delta}$ & Acc. & F1 & AUC & P. & R. & Avg.~$\boldsymbol{\Delta}$ \\
    \hline
    cBERT \citep{devlin2019bert} & 76.07{\color{grayv} \footnotesize $\pm$1.3} & 75.49{\color{grayv} \footnotesize $\pm$1.2} & 91.72{\color{grayv} \footnotesize $\pm$0.3} & 76.25{\color{grayv} \footnotesize $\pm$2.0} & 76.12{\color{grayv} \footnotesize $\pm$1.3} & - & 82.96{\color{grayv} \footnotesize $\pm$0.7} & 81.85{\color{grayv} \footnotesize $\pm$0.5} & 81.84{\color{grayv} \footnotesize $\pm$0.3} & 81.98{\color{grayv} \footnotesize $\pm$1.0} & 81.84{\color{grayv} \footnotesize $\pm$0.3} & - \\
    \rowcolor{lightgrayv} \quad + CGT (ours) & 79.29{\color{grayv} \footnotesize $\pm$1.8}$^*$ & 78.91{\color{grayv} \footnotesize $\pm$1.7}$^*$ & 93.24{\color{grayv} \footnotesize $\pm$0.3}$^*$ & 79.85{\color{grayv} \footnotesize $\pm$1.4}$^*$ & 79.35{\color{grayv} \footnotesize $\pm$1.8}$^*$ & \textbf{+3.00} & 84.94{\color{grayv} \footnotesize $\pm$0.6}$^*$ & 83.87{\color{grayv} \footnotesize $\pm$0.7}$^*$ & 83.67{\color{grayv} \footnotesize $\pm$1.0}$^*$ & 84.18{\color{grayv} \footnotesize $\pm$0.7}$^*$ & 83.67{\color{grayv} \footnotesize $\pm$1.0}$^*$ & \textbf{+1.97} \\
    
    dEFEND \citep{shu2019defend} & 75.72{\color{grayv} \footnotesize $\pm$2.0} & 75.17{\color{grayv} \footnotesize $\pm$2.2} & 91.44{\color{grayv} \footnotesize $\pm$0.7} & 75.62{\color{grayv} \footnotesize $\pm$2.0} & 75.82{\color{grayv} \footnotesize $\pm$1.6} & - & 82.80{\color{grayv} \footnotesize $\pm$1.0} & 81.98{\color{grayv} \footnotesize $\pm$0.8} & 82.60{\color{grayv} \footnotesize $\pm$0.8} & 81.87{\color{grayv} \footnotesize $\pm$0.8} & 82.60{\color{grayv} \footnotesize $\pm$0.8} & - \\
    \rowcolor{lightgrayv} \quad + CGT (ours) & 79.82{\color{grayv} \footnotesize $\pm$1.5}$^*$ & 79.27{\color{grayv} \footnotesize $\pm$1.8}$^*$ & 92.42{\color{grayv} \footnotesize $\pm$0.9}$^*$ & 79.97{\color{grayv} \footnotesize $\pm$1.9}$^*$ & 79.80{\color{grayv} \footnotesize $\pm$1.5}$^*$ & \textbf{+3.50} & 84.22{\color{grayv} \footnotesize $\pm$0.9}$^*$ & 83.35{\color{grayv} \footnotesize $\pm$0.8}$^*$ & 83.63{\color{grayv} \footnotesize $\pm$0.8}$^*$ & 83.26{\color{grayv} \footnotesize $\pm$1.0}$^*$ & 83.63{\color{grayv} \footnotesize $\pm$0.8}$^*$ & \textbf{+1.25} \\
    
    BERTEmo \citep{zhang2021mining} & 75.90{\color{grayv} \footnotesize $\pm$1.9} & 75.39{\color{grayv} \footnotesize $\pm$2.0} & 91.63{\color{grayv} \footnotesize $\pm$0.4} & 76.02{\color{grayv} \footnotesize $\pm$1.6} & 75.87{\color{grayv} \footnotesize $\pm$1.5} & - & 82.75{\color{grayv} \footnotesize $\pm$0.9} & 81.72{\color{grayv} \footnotesize $\pm$0.9} & 81.85{\color{grayv} \footnotesize $\pm$0.8} & 81.62{\color{grayv} \footnotesize $\pm$0.9} & 81.85{\color{grayv} \footnotesize $\pm$0.8} & - \\
    \rowcolor{lightgrayv} \quad + CGT (ours) & 78.57{\color{grayv} \footnotesize $\pm$1.9}$^*$ & 77.73{\color{grayv} \footnotesize $\pm$2.3}$^*$ & 93.25{\color{grayv} \footnotesize $\pm$0.7}$^*$ & 79.63{\color{grayv} \footnotesize $\pm$1.7}$^*$ & 78.67{\color{grayv} \footnotesize $\pm$1.8}$^*$ & \textbf{+2.61} & 84.72{\color{grayv} \footnotesize $\pm$1.1}$^*$ & 83.68{\color{grayv} \footnotesize $\pm$0.7}$^*$ & 83.52{\color{grayv} \footnotesize $\pm$0.7}$^*$ & 83.87{\color{grayv} \footnotesize $\pm$0.8}$^*$ & 83.52{\color{grayv} \footnotesize $\pm$0.7}$^*$ &  \textbf{+1.90} \\
    
    KAHAN \citep{tseng2022kahan} & 75.89{\color{grayv} \footnotesize $\pm$2.0} & 75.70{\color{grayv} \footnotesize $\pm$1.9} & 92.58{\color{grayv} \footnotesize $\pm$0.3} & 76.21{\color{grayv} \footnotesize $\pm$1.8} & 75.91{\color{grayv} \footnotesize $\pm$2.0} & - & 82.93{\color{grayv} \footnotesize $\pm$1.1} & 81.83{\color{grayv} \footnotesize $\pm$0.8} & 81.87{\color{grayv} \footnotesize $\pm$1.1} & 81.95{\color{grayv} \footnotesize $\pm$1.0} & 81.87{\color{grayv} \footnotesize $\pm$1.1} & - \\
    \rowcolor{lightgrayv} \quad + CGT (ours) & 78.57{\color{grayv} \footnotesize $\pm$1.0}$^*$ & 78.15{\color{grayv} \footnotesize $\pm$1.0}$^*$ & 92.11{\color{grayv} \footnotesize $\pm$0.5} & 79.77{\color{grayv} \footnotesize $\pm$1.7}$^*$ & 78.57{\color{grayv} \footnotesize $\pm$1.0}$^*$ & \textbf{+2.18} & 84.38{\color{grayv} \footnotesize $\pm$1.1}$^*$ & 83.45{\color{grayv} \footnotesize $\pm$1.0}$^*$ & 83.57{\color{grayv} \footnotesize $\pm$1.1}$^*$ & 83.46{\color{grayv} \footnotesize $\pm$0.9}$^*$ & 83.57{\color{grayv} \footnotesize $\pm$1.1}$^*$ & \textbf{+1.60} \\

    CAS-FEND \citep{nan2024exploiting} & 75.18{\color{grayv} \footnotesize $\pm$1.2} & 74.99{\color{grayv} \footnotesize $\pm$1.2} & 91.56{\color{grayv} \footnotesize $\pm$0.6} & 75.13{\color{grayv} \footnotesize $\pm$1.1} & 75.20{\color{grayv} \footnotesize $\pm$1.3} & - & 83.25{\color{grayv} \footnotesize $\pm$0.6} & 81.69{\color{grayv} \footnotesize $\pm$0.6} & 80.99{\color{grayv} \footnotesize $\pm$0.8} & 83.00{\color{grayv} \footnotesize $\pm$1.1} & 80.99{\color{grayv} \footnotesize $\pm$0.8} & - \\
    \rowcolor{lightgrayv} \quad + CGT (ours) & 78.93{\color{grayv} \footnotesize $\pm$1.8}$^*$ & 78.86{\color{grayv} \footnotesize $\pm$1.8}$^*$ & 92.31{\color{grayv} \footnotesize $\pm$0.5}$^*$ & 79.39{\color{grayv} \footnotesize $\pm$1.5}$^*$ & 78.89{\color{grayv} \footnotesize $\pm$1.3}$^*$ & \textbf{+3.26} & 84.54{\color{grayv} \footnotesize $\pm$0.7}$^*$ & 83.33{\color{grayv} \footnotesize $\pm$0.8}$^*$ & 82.93{\color{grayv} \footnotesize $\pm$0.9}$^*$ & 83.95{\color{grayv} \footnotesize $\pm$0.9}$^*$ & 82.93{\color{grayv} \footnotesize $\pm$0.9}$^*$ & \textbf{+1.55} \\
    
    \hline
    \textbf{\baby} & 76.43{\color{grayv} \footnotesize $\pm$1.6} & 76.18{\color{grayv} \footnotesize $\pm$1.4} & 91.92{\color{grayv} \footnotesize $\pm$0.3} & 77.12{\color{grayv} \footnotesize $\pm$1.5} & 76.47{\color{grayv} \footnotesize $\pm$1.7} & - & 83.72{\color{grayv} \footnotesize $\pm$0.3} & 82.66{\color{grayv} \footnotesize $\pm$0.4} & 82.60{\color{grayv} \footnotesize $\pm$0.4} & 82.72{\color{grayv} \footnotesize $\pm$0.4} & 82.60{\color{grayv} \footnotesize $\pm$0.4} & - \\
    \rowcolor{lightgrayv} \quad + CGT (ours) & 80.36{\color{grayv} \footnotesize $\pm$1.3}$^*$ & 80.11{\color{grayv} \footnotesize $\pm$1.4}$^*$ & 93.54{\color{grayv} \footnotesize $\pm$1.0}$^*$ & 80.68{\color{grayv} \footnotesize $\pm$1.6}$^*$ & 80.28{\color{grayv} \footnotesize $\pm$1.5}$^*$ & \textbf{+3.37} & 86.21{\color{grayv} \footnotesize $\pm$1.0}$^*$ & 85.32{\color{grayv} \footnotesize $\pm$1.0}$^*$ & 85.29{\color{grayv} \footnotesize $\pm$1.1}$^*$ & 85.48{\color{grayv} \footnotesize $\pm$1.2}$^*$ & 85.29{\color{grayv} \footnotesize $\pm$1.1}$^*$ & \textbf{+2.66} \\
    \hline
    \specialrule{0em}{0.5pt}{0.5pt}
    \hline
    
    \multirow{2}{*}{\quad \quad Model} & \multicolumn{6}{c}{Dataset: \textit{Twitter16} \citep{ma2017detect}} & \multicolumn{6}{c}{Dataset: \textit{Weibo20} \citep{zhang2021mining}} \\
    
    \cmidrule(r){2-7} \cmidrule(r){8-13} 
    & Acc. & F1 & AUC & P. & R. & Avg.~$\boldsymbol{\Delta}$ & Acc. & F1 & AUC & P. & R. & Avg.~$\boldsymbol{\Delta}$ \\
    \hline
    cBERT \citep{devlin2019bert} & 74.54{\color{grayv} \footnotesize $\pm$2.5} & 74.21{\color{grayv} \footnotesize $\pm$2.0} & 92.27{\color{grayv} \footnotesize $\pm$1.8} & 75.50{\color{grayv} \footnotesize $\pm$2.7} & 74.54{\color{grayv} \footnotesize $\pm$2.5} & - & 86.16{\color{grayv} \footnotesize $\pm$0.6} & 86.14{\color{grayv} \footnotesize $\pm$0.6} & 86.19{\color{grayv} \footnotesize $\pm$0.6} & 86.47{\color{grayv} \footnotesize $\pm$0.6} & 86.19{\color{grayv} \footnotesize $\pm$0.6} & - \\
    
    \rowcolor{lightgrayv} \quad + CGT (ours) & 78.44{\color{grayv} \footnotesize $\pm$1.6}$^*$ & 78.40{\color{grayv} \footnotesize $\pm$1.6}$^*$ & 93.77{\color{grayv} \footnotesize $\pm$0.7}$^*$ & 79.52{\color{grayv} \footnotesize $\pm$1.2}$^*$ & 78.85{\color{grayv} \footnotesize $\pm$1.7}$^*$ & \textbf{+3.58} & 88.33{\color{grayv} \footnotesize $\pm$0.8}$^*$ & 88.32{\color{grayv} \footnotesize $\pm$0.8}$^*$ & 88.32{\color{grayv} \footnotesize $\pm$0.8}$^*$ & 88.38{\color{grayv} \footnotesize $\pm$0.8}$^*$ & 88.32{\color{grayv} \footnotesize $\pm$0.8}$^*$ & \textbf{+2.10} \\
    
    dEFEND \citep{shu2019defend} & 72.98{\color{grayv} \footnotesize $\pm$1.9} & 72.82{\color{grayv} \footnotesize $\pm$2.1} & 92.50{\color{grayv} \footnotesize $\pm$0.6} & 75.25{\color{grayv} \footnotesize $\pm$1.9} & 72.96{\color{grayv} \footnotesize $\pm$2.0} & - & 86.28{\color{grayv} \footnotesize $\pm$0.5} & 86.26{\color{grayv} \footnotesize $\pm$0.5} & 86.27{\color{grayv} \footnotesize $\pm$0.5} & 86.36{\color{grayv} \footnotesize $\pm$0.4} & 86.27{\color{grayv} \footnotesize $\pm$0.5} & - \\
    
    \rowcolor{lightgrayv} \quad + CGT (ours) & 77.66{\color{grayv} \footnotesize $\pm$1.5}$^*$ & 77.65{\color{grayv} \footnotesize $\pm$1.6}$^*$ & 93.72{\color{grayv} \footnotesize $\pm$0.3}$^*$ & 79.00{\color{grayv} \footnotesize $\pm$1.5}$^*$ & 78.15{\color{grayv} \footnotesize $\pm$1.4}$^*$ & \textbf{+3.93} & 88.38{\color{grayv} \footnotesize $\pm$0.8}$^*$ & 88.38{\color{grayv} \footnotesize $\pm$0.8}$^*$ & 88.39{\color{grayv} \footnotesize $\pm$0.8}$^*$ & 88.41{\color{grayv} \footnotesize $\pm$0.8}$^*$ & 88.39{\color{grayv} \footnotesize $\pm$0.8}$^*$ & \textbf{+2.10} \\
    
    BERTEmo \citep{zhang2021mining} & 74.02{\color{grayv} \footnotesize $\pm$2.4} & 73.83{\color{grayv} \footnotesize $\pm$2.4} & 92.23{\color{grayv} \footnotesize $\pm$2.1} & 75.16{\color{grayv} \footnotesize $\pm$1.9} & 74.26{\color{grayv} \footnotesize $\pm$2.6} & - & 86.03{\color{grayv} \footnotesize $\pm$0.9} & 86.00{\color{grayv} \footnotesize $\pm$0.9} & 86.05{\color{grayv} \footnotesize $\pm$0.9} & 86.33{\color{grayv} \footnotesize $\pm$0.8} & 86.05{\color{grayv} \footnotesize $\pm$0.9} & - \\
    
    \rowcolor{lightgrayv} \quad + CGT (ours) & 77.14{\color{grayv} \footnotesize $\pm$1.7}$^*$ & 77.12{\color{grayv} \footnotesize $\pm$1.7}$^*$ & 92.87{\color{grayv} \footnotesize $\pm$1.8} & 77.87{\color{grayv} \footnotesize $\pm$2.2}$^*$ & 77.35{\color{grayv} \footnotesize $\pm$1.8}$^*$ & \textbf{+2.57} & 88.08{\color{grayv} \footnotesize $\pm$0.7}$^*$ & 88.08{\color{grayv} \footnotesize $\pm$0.7}$^*$ & 88.09{\color{grayv} \footnotesize $\pm$0.7}$^*$ & 88.13{\color{grayv} \footnotesize $\pm$0.7}$^*$ & 88.09{\color{grayv} \footnotesize $\pm$0.7}$^*$ & \textbf{+2.00} \\
    
    KAHAN \citep{tseng2022kahan} & 74.80{\color{grayv} \footnotesize $\pm$1.9} & 74.89{\color{grayv} \footnotesize $\pm$2.0} & 91.04{\color{grayv} \footnotesize $\pm$1.2} & 75.41{\color{grayv} \footnotesize $\pm$2.0} & 74.93{\color{grayv} \footnotesize $\pm$2.0} & - & 86.13{\color{grayv} \footnotesize $\pm$0.3} & 86.13{\color{grayv} \footnotesize $\pm$0.3} & 86.14{\color{grayv} \footnotesize $\pm$0.3} & 86.20{\color{grayv} \footnotesize $\pm$0.3} & 86.14{\color{grayv} \footnotesize $\pm$0.3} & - \\
    
    \rowcolor{lightgrayv} \quad + CGT (ours) & 77.92{\color{grayv} \footnotesize $\pm$0.9}$^*$ & 77.98{\color{grayv} \footnotesize $\pm$0.9}$^*$ & 92.28{\color{grayv} \footnotesize $\pm$0.3}$^*$ & 78.21{\color{grayv} \footnotesize $\pm$1.0}$^*$ & 78.11{\color{grayv} \footnotesize $\pm$0.9}$^*$ & \textbf{+2.68} & 88.25{\color{grayv} \footnotesize $\pm$0.5}$^*$ & 88.25{\color{grayv} \footnotesize $\pm$0.5}$^*$ & 88.26{\color{grayv} \footnotesize $\pm$0.5}$^*$ & 88.27{\color{grayv} \footnotesize $\pm$0.5}$^*$ & 88.26{\color{grayv} \footnotesize $\pm$0.5}$^*$ & \textbf{+2.11} \\

    CAS-FEND \citep{nan2024exploiting} & 73.76{\color{grayv} \footnotesize $\pm$1.8} & 73.69{\color{grayv} \footnotesize $\pm$2.0} & 91.44{\color{grayv} \footnotesize $\pm$0.9} & 74.32{\color{grayv} \footnotesize $\pm$1.8} & 73.97{\color{grayv} \footnotesize $\pm$1.8} & - & 85.51{\color{grayv} \footnotesize $\pm$0.6} & 85.47{\color{grayv} \footnotesize $\pm$0.6} & 85.54{\color{grayv} \footnotesize $\pm$0.6} & 85.89{\color{grayv} \footnotesize $\pm$0.6} & 85.54{\color{grayv} \footnotesize $\pm$0.6} & - \\
    
    \rowcolor{lightgrayv} \quad + CGT (ours) & 78.44{\color{grayv} \footnotesize $\pm$1.6}$^*$ & 78.32{\color{grayv} \footnotesize $\pm$1.7}$^*$ & 93.90{\color{grayv} \footnotesize $\pm$1.2}$^*$ & 78.81{\color{grayv} \footnotesize $\pm$1.6}$^*$ & 79.01{\color{grayv} \footnotesize $\pm$1.6}$^*$ & \textbf{+4.26} & 87.92{\color{grayv} \footnotesize $\pm$0.8}$^*$ & 87.92{\color{grayv} \footnotesize $\pm$0.8}$^*$ & 87.93{\color{grayv} \footnotesize $\pm$0.8}$^*$ & 87.94{\color{grayv} \footnotesize $\pm$0.8}$^*$ & 87.93{\color{grayv} \footnotesize $\pm$0.8}$^*$ & \textbf{+2.34} \\
    
    \hline
    \textbf{\baby} & 75.06{\color{grayv} \footnotesize $\pm$1.9} & 74.98{\color{grayv} \footnotesize $\pm$1.9} & 92.92{\color{grayv} \footnotesize $\pm$1.1} & 76.36{\color{grayv} \footnotesize $\pm$2.0} & 75.48{\color{grayv} \footnotesize $\pm$1.7} & - & 86.56{\color{grayv} \footnotesize $\pm$0.9} & 86.55{\color{grayv} \footnotesize $\pm$0.9} & 86.56{\color{grayv} \footnotesize $\pm$0.9} & 86.58{\color{grayv} \footnotesize $\pm$0.9} & 86.56{\color{grayv} \footnotesize $\pm$0.9} & - \\
    
    \rowcolor{lightgrayv} \quad + CGT (ours) & 79.55{\color{grayv} \footnotesize $\pm$1.2}$^*$ & 79.53{\color{grayv} \footnotesize $\pm$1.3}$^*$ & 93.69{\color{grayv} \footnotesize $\pm$0.5}$^*$ & 80.05{\color{grayv} \footnotesize $\pm$1.1}$^*$ & 79.82{\color{grayv} \footnotesize $\pm$1.2}$^*$ & \textbf{+3.57} & 88.63{\color{grayv} \footnotesize $\pm$0.5}$^*$ & 88.63{\color{grayv} \footnotesize $\pm$0.5}$^*$ & 88.65{\color{grayv} \footnotesize $\pm$0.5}$^*$ & 88.72{\color{grayv} \footnotesize $\pm$0.4}$^*$ & 88.65{\color{grayv} \footnotesize $\pm$0.5}$^*$ & \textbf{+2.09} \\
    \bottomrule
  \end{tabular} }
\end{table*}

In this section, we conduct experiments on four prevalent RED datasets to validate the performance of \baby.

\subsection{Experimental Settings}

\vspace{3pt} \noindent
\textbf{Datasets.}
We evaluate the proposed \baby method using the following four public RD datasets.
\textbf{\textit{Twitter15}} \citep{ma2017detect} and \textbf{\textit{Twitter16}} \citep{ma2017detect} were collected from the Twitter platform and labeled using rumor debunking websites, \eg \textit{snopes.com}, with four labels: non-rumor, false rumor, true rumor, and unverified rumor. 
\textbf{\textit{Weibo16}} \citep{ma2016detecting} and \textbf{\textit{Weibo20}} \citep{zhang2021mining} are Chinese RD datasets collected from the Chinese social platform Weibo. 
For clarity, the statistics of these datasets are presented in Table~\ref{dataset}.

\begin{table*}[t]
\centering
\renewcommand\arraystretch{0.93}
  \caption{Experimental results under the RED scenario (we fix $M = 2$ and $M^\prime = 2$).}
  \label{resultcontent}
  \small
  \setlength{\tabcolsep}{5pt}{
  \begin{tabular}{m{1.95cm}m{0.94cm}<{\centering}m{0.94cm}<{\centering}m{0.94cm}<{\centering}m{0.94cm}<{\centering}m{0.94cm}<{\centering}m{0.80cm}<{\centering}m{0.94cm}<{\centering}m{0.94cm}<{\centering}m{0.94cm}<{\centering}m{0.94cm}<{\centering}m{0.94cm}<{\centering}m{0.80cm}<{\centering}}
    \toprule
    \multirow{2}{*}{\quad \quad Model} & \multicolumn{6}{c}{Dataset: \textit{Twitter15} \citep{ma2017detect}} & \multicolumn{6}{c}{Dataset: \textit{Weibo16} \citep{ma2016detecting}} \\
    
    \cmidrule(r){2-7} \cmidrule(r){8-13} 
    & Acc. & F1 & AUC & P. & R. & Avg.~$\boldsymbol{\Delta}$ & Acc. & F1 & AUC & P. & R. & Avg.~$\boldsymbol{\Delta}$ \\
    \hline
    cBERT \citep{devlin2019bert} & 76.61{\color{grayv} \footnotesize $\pm$1.4} & 76.02{\color{grayv} \footnotesize $\pm$1.5} & 91.86{\color{grayv} \footnotesize $\pm$1.0} & 76.83{\color{grayv} \footnotesize $\pm$1.3} & 76.64{\color{grayv} \footnotesize $\pm$1.5} & - & 82.56{\color{grayv} \footnotesize $\pm$0.5} & 81.23{\color{grayv} \footnotesize $\pm$0.7} & 80.97{\color{grayv} \footnotesize $\pm$1.2} & 81.79{\color{grayv} \footnotesize $\pm$0.6} & 80.97{\color{grayv} \footnotesize $\pm$1.2} & - \\
    \rowcolor{lightgrayv} \quad + CGT (ours) & 78.22{\color{grayv} \footnotesize $\pm$1.5}$^*$ & 77.81{\color{grayv} \footnotesize $\pm$1.2}$^*$ & 91.88{\color{grayv} \footnotesize $\pm$0.3} & 78.99{\color{grayv} \footnotesize $\pm$1.2}$^*$ & 78.25{\color{grayv} \footnotesize $\pm$1.6}$^*$ & \textbf{+1.44} & 84.88{\color{grayv} \footnotesize $\pm$1.0}$^*$ & 83.91{\color{grayv} \footnotesize $\pm$0.9}$^*$ & 83.91{\color{grayv} \footnotesize $\pm$1.1}$^*$ & 84.21{\color{grayv} \footnotesize $\pm$1.3}$^*$ & 83.91{\color{grayv} \footnotesize $\pm$1.1}$^*$ & \textbf{+2.66} \\
    
    dEFEND \citep{shu2019defend} & 76.79{\color{grayv} \footnotesize $\pm$1.2} & 75.77{\color{grayv} \footnotesize $\pm$1.3} & 92.12{\color{grayv} \footnotesize $\pm$0.8} & 78.39{\color{grayv} \footnotesize $\pm$1.0} & 76.96{\color{grayv} \footnotesize $\pm$1.1} & - & 82.58{\color{grayv} \footnotesize $\pm$1.2} & 81.31{\color{grayv} \footnotesize $\pm$1.1} & 81.03{\color{grayv} \footnotesize $\pm$1.2} & 81.75{\color{grayv} \footnotesize $\pm$0.8} & 81.03{\color{grayv} \footnotesize $\pm$1.2} & - \\
    \rowcolor{lightgrayv} \quad + CGT (ours) & 79.42{\color{grayv} \footnotesize $\pm$0.9}$^*$ & 79.08{\color{grayv} \footnotesize $\pm$1.2}$^*$ & 93.01{\color{grayv} \footnotesize $\pm$0.5}$^*$ & 80.34{\color{grayv} \footnotesize $\pm$1.1}$^*$ & 79.56{\color{grayv} \footnotesize $\pm$0.9}$^*$ & \textbf{+2.28} & 84.91{\color{grayv} \footnotesize $\pm$0.4}$^*$ & 83.66{\color{grayv} \footnotesize $\pm$0.3}$^*$ & 83.13{\color{grayv} \footnotesize $\pm$0.4}$^*$ & 84.50{\color{grayv} \footnotesize $\pm$0.8}$^*$ & 83.13{\color{grayv} \footnotesize $\pm$0.4}$^*$ & \textbf{+2.33} \\
    
    BERTEmo \citep{zhang2021mining} & 76.43{\color{grayv} \footnotesize $\pm$1.0} & 75.70{\color{grayv} \footnotesize $\pm$1.2} & 92.54{\color{grayv} \footnotesize $\pm$0.6} & 77.76{\color{grayv} \footnotesize $\pm$1.4} & 76.61{\color{grayv} \footnotesize $\pm$1.5} & - & 82.69{\color{grayv} \footnotesize $\pm$1.3} & 81.60{\color{grayv} \footnotesize $\pm$1.1} & 81.64{\color{grayv} \footnotesize $\pm$0.8} & 81.79{\color{grayv} \footnotesize $\pm$1.1} & 81.64{\color{grayv} \footnotesize $\pm$0.8} & - \\
    \rowcolor{lightgrayv} \quad + CGT (ours) & 78.57{\color{grayv} \footnotesize $\pm$0.9}$^*$ & 77.73{\color{grayv} \footnotesize $\pm$0.9}$^*$ & 93.25{\color{grayv} \footnotesize $\pm$0.7}$^*$ & 79.63{\color{grayv} \footnotesize $\pm$1.2}$^*$ & 78.67{\color{grayv} \footnotesize $\pm$1.3}$^*$ & \textbf{+1.76} & 84.38{\color{grayv} \footnotesize $\pm$0.6}$^*$ & 83.28{\color{grayv} \footnotesize $\pm$0.8}$^*$ & 83.14{\color{grayv} \footnotesize $\pm$0.9}$^*$ & 83.52{\color{grayv} \footnotesize $\pm$0.7}$^*$ & 83.14{\color{grayv} \footnotesize $\pm$0.9}$^*$ & \textbf{+1.62} \\
    
    KAHAN \citep{tseng2022kahan} & 76.43{\color{grayv} \footnotesize $\pm$1.2} & 75.89{\color{grayv} \footnotesize $\pm$1.2} & 92.56{\color{grayv} \footnotesize $\pm$0.7} & 76.40{\color{grayv} \footnotesize $\pm$1.1} & 76.42{\color{grayv} \footnotesize $\pm$1.2} & - & 82.51{\color{grayv} \footnotesize $\pm$1.1} & 81.24{\color{grayv} \footnotesize $\pm$1.0} & 81.01{\color{grayv} \footnotesize $\pm$0.8} & 81.64{\color{grayv} \footnotesize $\pm$1.0} & 81.01{\color{grayv} \footnotesize $\pm$0.8} & - \\
    \rowcolor{lightgrayv} \quad + CGT (ours) & 79.29{\color{grayv} \footnotesize $\pm$1.1}$^*$ & 78.66{\color{grayv} \footnotesize $\pm$0.9}$^*$ & 93.17{\color{grayv} \footnotesize $\pm$0.6}$^*$ & 80.02{\color{grayv} \footnotesize $\pm$1.2}$^*$ & 79.38{\color{grayv} \footnotesize $\pm$1.1}$^*$ & \textbf{+2.56} & 84.41{\color{grayv} \footnotesize $\pm$0.7}$^*$ & 83.46{\color{grayv} \footnotesize $\pm$0.7}$^*$ & 83.55{\color{grayv} \footnotesize $\pm$0.9}$^*$ & 83.46{\color{grayv} \footnotesize $\pm$0.8}$^*$ & 83.55{\color{grayv} \footnotesize $\pm$0.9}$^*$ & \textbf{+2.20} \\

    CAS-FEND \citep{nan2024exploiting} & 75.54{\color{grayv} \footnotesize $\pm$0.8} & 75.33{\color{grayv} \footnotesize $\pm$0.8} & 91.39{\color{grayv} \footnotesize $\pm$0.7} & 75.68{\color{grayv} \footnotesize $\pm$0.8} & 75.45{\color{grayv} \footnotesize $\pm$0.7} & - & 83.19{\color{grayv} \footnotesize $\pm$1.1} & 81.99{\color{grayv} \footnotesize $\pm$1.0}  & 81.77{\color{grayv} \footnotesize $\pm$0.7} & 82.37{\color{grayv} \footnotesize $\pm$1.6} & 81.77{\color{grayv} \footnotesize $\pm$0.7} & - \\
    \rowcolor{lightgrayv} \quad + CGT (ours) & 78.93{\color{grayv} \footnotesize $\pm$0.9}$^*$ & 78.66{\color{grayv} \footnotesize $\pm$0.9}$^*$ & 92.66{\color{grayv} \footnotesize $\pm$0.7}$^*$ & 79.16{\color{grayv} \footnotesize $\pm$1.0}$^*$ & 78.90{\color{grayv} \footnotesize $\pm$0.8}$^*$ & \textbf{+2.98} & 84.51{\color{grayv} \footnotesize $\pm$0.8}$^*$ & 82.95{\color{grayv} \footnotesize $\pm$1.0}$^*$ & 82.64{\color{grayv} \footnotesize $\pm$1.0}$^*$ & 83.90{\color{grayv} \footnotesize $\pm$0.9}$^*$ & 82.64{\color{grayv} \footnotesize $\pm$1.0}$^*$ & \textbf{+1.11} \\
    
    \hline
    \textbf{\baby} & 76.79{\color{grayv} \footnotesize $\pm$1.5} & 76.30{\color{grayv} \footnotesize $\pm$1.2} & 91.79{\color{grayv} \footnotesize $\pm$0.0} & 77.98{\color{grayv} \footnotesize $\pm$1.8} & 76.93{\color{grayv} \footnotesize $\pm$1.1} & - & 82.23{\color{grayv} \footnotesize $\pm$0.7} & 82.00{\color{grayv} \footnotesize $\pm$0.5} & 81.76{\color{grayv} \footnotesize $\pm$0.7} & 82.60{\color{grayv} \footnotesize $\pm$1.1} & 81.76{\color{grayv} \footnotesize $\pm$0.7} & - \\
    \rowcolor{lightgrayv} \quad + CGT (ours) & 80.00{\color{grayv} \footnotesize $\pm$1.3}$^*$ & 79.46{\color{grayv} \footnotesize $\pm$1.3}$^*$ & 92.75{\color{grayv} \footnotesize $\pm$0.8}$^*$ & 81.30{\color{grayv} \footnotesize $\pm$1.2}$^*$ & 79.98{\color{grayv} \footnotesize $\pm$1.2}$^*$ & \textbf{+2.74} & 86.00{\color{grayv} \footnotesize $\pm$1.1}$^*$ & 85.13{\color{grayv} \footnotesize $\pm$1.4}$^*$ & 85.18{\color{grayv} \footnotesize $\pm$1.5}$^*$ & 85.08{\color{grayv} \footnotesize $\pm$1.4}$^*$ & 85.18{\color{grayv} \footnotesize $\pm$1.5}$^*$ & \textbf{+3.24} \\
    \hline
    \specialrule{0em}{0.5pt}{0.5pt}
    \hline
    
    \multirow{2}{*}{\quad \quad Model} & \multicolumn{6}{c}{Dataset: \textit{Twitter16} \citep{ma2017detect}} & \multicolumn{6}{c}{Dataset: \textit{Weibo20} \citep{zhang2021mining}} \\
    
    \cmidrule(r){2-7} \cmidrule(r){8-13} 
    & Acc. & F1 & AUC & P. & R. & Avg.~$\boldsymbol{\Delta}$ & Acc. & F1 & AUC & P. & R. & Avg.~$\boldsymbol{\Delta}$ \\
    \hline
    cBERT \citep{devlin2019bert} & 75.32{\color{grayv} \footnotesize $\pm$2.4} & 75.07{\color{grayv} \footnotesize $\pm$2.5} & 93.18{\color{grayv} \footnotesize $\pm$0.9} & 76.90{\color{grayv} \footnotesize $\pm$2.1} & 75.67{\color{grayv} \footnotesize $\pm$2.4} & - & 85.84{\color{grayv} \footnotesize $\pm$0.4} & 85.82{\color{grayv} \footnotesize $\pm$0.4} & 85.86{\color{grayv} \footnotesize $\pm$0.4} & 86.01{\color{grayv} \footnotesize $\pm$0.3} & 85.86{\color{grayv} \footnotesize $\pm$0.4} & - \\
    
    \rowcolor{lightgrayv} \quad + CGT (ours) & 77.92{\color{grayv} \footnotesize $\pm$2.0}$^*$ & 77.96{\color{grayv} \footnotesize $\pm$2.1}$^*$ & 94.07{\color{grayv} \footnotesize $\pm$0.3}$^*$ & 78.53{\color{grayv} \footnotesize $\pm$2.2}$^*$ & 78.26{\color{grayv} \footnotesize $\pm$2.1}$^*$ & \textbf{+2.12} & 87.78{\color{grayv} \footnotesize $\pm$0.3}$^*$ & 87.78{\color{grayv} \footnotesize $\pm$0.3}$^*$ & 87.79{\color{grayv} \footnotesize $\pm$0.3}$^*${\color{grayv} \footnotesize $\pm$0.3}$^*$ & 87.85{\color{grayv} \footnotesize $\pm$0.3}$^*$ & 87.79{\color{grayv} \footnotesize $\pm$0.3}$^*$ & \textbf{+1.92} \\
    
    dEFEND \citep{shu2019defend} & 75.06{\color{grayv} \footnotesize $\pm$2.1} & 75.09{\color{grayv} \footnotesize $\pm$2.0} & 93.31{\color{grayv} \footnotesize $\pm$0.5} & 76.49{\color{grayv} \footnotesize $\pm$1.4} & 75.34{\color{grayv} \footnotesize $\pm$2.0} & - & 85.87{\color{grayv} \footnotesize $\pm$0.7} & 85.86{\color{grayv} \footnotesize $\pm$0.7} & 85.88{\color{grayv} \footnotesize $\pm$0.7} & 85.97{\color{grayv} \footnotesize $\pm$0.7} & 85.88{\color{grayv} \footnotesize $\pm$0.7} & - \\
    
    \rowcolor{lightgrayv} \quad + CGT (ours) & 77.92{\color{grayv} \footnotesize $\pm$2.4}$^*$ & 77.86{\color{grayv} \footnotesize $\pm$2.4}$^*$ & 94.32{\color{grayv} \footnotesize $\pm$1.0}$^*$ & 79.65{\color{grayv} \footnotesize $\pm$1.9}$^*$ & 78.34{\color{grayv} \footnotesize $\pm$2.0}$^*$ & \textbf{+2.56} & 87.94{\color{grayv} \footnotesize $\pm$0.7}$^*$ & 87.93{\color{grayv} \footnotesize $\pm$0.7}$^*$ & 87.95{\color{grayv} \footnotesize $\pm$0.7}$^*$ & 88.03{\color{grayv} \footnotesize $\pm$0.8}$^*$ & 87.95{\color{grayv} \footnotesize $\pm$0.7}$^*$ & \textbf{+2.07} \\
    
    BERTEmo \citep{zhang2021mining} & 74.80{\color{grayv} \footnotesize $\pm$1.9} & 74.78{\color{grayv} \footnotesize $\pm$1.9} & 92.22{\color{grayv} \footnotesize $\pm$1.3} & 75.83{\color{grayv} \footnotesize $\pm$2.3} & 74.95{\color{grayv} \footnotesize $\pm$2.0} & - & 85.62{\color{grayv} \footnotesize $\pm$1.0} & 85.60{\color{grayv} \footnotesize $\pm$1.0} & 85.64{\color{grayv} \footnotesize $\pm$1.0} & 85.80{\color{grayv} \footnotesize $\pm$0.9} & 85.64{\color{grayv} \footnotesize $\pm$1.0} & - \\
    
    \rowcolor{lightgrayv} \quad + CGT (ours) & 77.40{\color{grayv} \footnotesize $\pm$1.8}$^*$ & 77.48{\color{grayv} \footnotesize $\pm$1.8}$^*$ & 94.11{\color{grayv} \footnotesize $\pm$0.4}$^*$ & 78.85{\color{grayv} \footnotesize $\pm$1.7}$^*$ & 77.66{\color{grayv} \footnotesize $\pm$1.5}$^*$ & \textbf{+2.58} & 87.88{\color{grayv} \footnotesize $\pm$0.3}$^*$ & 87.87{\color{grayv} \footnotesize $\pm$0.3}$^*$ & 87.88{\color{grayv} \footnotesize $\pm$0.3}$^*$ & 87.92{\color{grayv} \footnotesize $\pm$0.3}$^*$ & 87.88{\color{grayv} \footnotesize $\pm$0.3}$^*$ & \textbf{+2.23} \\
    
    KAHAN \citep{tseng2022kahan} & 74.80{\color{grayv} \footnotesize $\pm$2.3} & 74.85{\color{grayv} \footnotesize $\pm$2.2} & 92.32{\color{grayv} \footnotesize $\pm$0.6} & 75.22{\color{grayv} \footnotesize $\pm$2.0} & 74.89{\color{grayv} \footnotesize $\pm$2.3} & - & 85.90{\color{grayv} \footnotesize $\pm$0.7} & 85.89{\color{grayv} \footnotesize $\pm$0.7} & 85.92{\color{grayv} \footnotesize $\pm$0.7} & 86.06{\color{grayv} \footnotesize $\pm$0.7} & 85.92{\color{grayv} \footnotesize $\pm$0.7} & - \\
    
    \rowcolor{lightgrayv} \quad + CGT (ours) & 78.25{\color{grayv} \footnotesize $\pm$1.6}$^*$ & 78.37{\color{grayv} \footnotesize $\pm$1.7}$^*$ & 92.44{\color{grayv} \footnotesize $\pm$0.4} & 78.55{\color{grayv} \footnotesize $\pm$1.5}$^*$ & 78.37{\color{grayv} \footnotesize $\pm$1.8}$^*$ & \textbf{+2.78} & 88.00{\color{grayv} \footnotesize $\pm$0.6}$^*$ & 87.99{\color{grayv} \footnotesize $\pm$0.6}$^*$ & 88.02{\color{grayv} \footnotesize $\pm$0.6}$^*$ & 88.18{\color{grayv} \footnotesize $\pm$0.5}$^*$ & 88.02{\color{grayv} \footnotesize $\pm$0.6}$^*$ & \textbf{+2.10} \\

    CAS-FEND \citep{nan2024exploiting} & 74.55{\color{grayv} \footnotesize $\pm$1.4} & 74.59{\color{grayv} \footnotesize $\pm$1.3} & 92.34{\color{grayv} \footnotesize $\pm$1.0} & 75.45{\color{grayv} \footnotesize $\pm$1.5} & 74.85{\color{grayv} \footnotesize $\pm$1.4} & - & 85.74{\color{grayv} \footnotesize $\pm$0.5} & 85.73{\color{grayv} \footnotesize $\pm$0.5} & 85.76{\color{grayv} \footnotesize $\pm$0.5} & 85.85{\color{grayv} \footnotesize $\pm$0.5} & 85.76{\color{grayv} \footnotesize $\pm$0.5} & \\
    
    \rowcolor{lightgrayv} \quad + CGT (ours) & 78.70{\color{grayv} \footnotesize $\pm$0.7}$^*$ & 78.71{\color{grayv} \footnotesize $\pm$0.7}$^*$ & 93.60{\color{grayv} \footnotesize $\pm$0.7}$^*$ & 78.78{\color{grayv} \footnotesize $\pm$0.6}$^*$ & 78.96{\color{grayv} \footnotesize $\pm$0.6}$^*$ & \textbf{+3.39} & 87.72{\color{grayv} \footnotesize $\pm$0.7}$^*$ & 87.72{\color{grayv} \footnotesize $\pm$0.7}$^*$ & 87.73{\color{grayv} \footnotesize $\pm$0.7}$^*$ & 87.80{\color{grayv} \footnotesize $\pm$0.6}$^*$ & 87.73{\color{grayv} \footnotesize $\pm$0.7}$^*$ & \textbf{+1.89} \\
    
    \hline
    \textbf{\baby} & 75.84{\color{grayv} \footnotesize $\pm$1.9} & 75.93{\color{grayv} \footnotesize $\pm$2.0} & 93.30{\color{grayv} \footnotesize $\pm$1.1} & 77.05{\color{grayv} \footnotesize $\pm$1.5} & 76.33{\color{grayv} \footnotesize $\pm$1.8} & - & 85.90{\color{grayv} \footnotesize $\pm$0.6} & 85.90{\color{grayv} \footnotesize $\pm$0.6} & 85.91{\color{grayv} \footnotesize $\pm$0.6} & 85.93{\color{grayv} \footnotesize $\pm$0.6} & 85.91{\color{grayv} \footnotesize $\pm$0.6} & - \\
    
    \rowcolor{lightgrayv} \quad + CGT (ours) & 79.48{\color{grayv} \footnotesize $\pm$2.1}$^*$ & 79.52{\color{grayv} \footnotesize $\pm$2.1}$^*$ & 94.64{\color{grayv} \footnotesize $\pm$0.7}$^*$ & 80.77{\color{grayv} \footnotesize $\pm$2.0}$^*$ & 79.86{\color{grayv} \footnotesize $\pm$2.1}$^*$ & \textbf{+3.16} & 88.14{\color{grayv} \footnotesize $\pm$0.5}$^*$ & 88.13{\color{grayv} \footnotesize $\pm$0.5}$^*$ & 88.16{\color{grayv} \footnotesize $\pm$0.5}$^*$ & 88.25{\color{grayv} \footnotesize $\pm$0.5}$^*$ & 88.16{\color{grayv} \footnotesize $\pm$0.5}$^*$ & \textbf{+2.26} \\
    \bottomrule
  \end{tabular} }
\end{table*}

\vspace{3pt} \noindent
\textbf{Baselines.}
To evaluate the performance of our generated comments and proposed method \baby, we compare the improvement of CGT and \baby on the following five baseline models: \textbf{cBERT} \citep{devlin2019bert}, \textbf{dEFEND} \citep{shu2019defend}, \textbf{BERTEmo} \citep{zhang2021mining}, \textbf{KAHAN} \citep{tseng2022kahan}, and \textbf{CAS-FEND} \citep{nan2024exploiting}.
We re-produce all baselines and use BERT as the backbone text encoder. In addition to comparing the improvements brought by generated comments and \baby into these baseline models, we also compare our generated comments to several methods that utilize LLMs for comment generation, \eg \textbf{DELL} \citep{wan2024dell} and \textbf{GenFEND} \citep{nan2024let}.
To be fair, we re-produce their comment generation process using the same Llama model that we use.

\vspace{3pt} \noindent
\textbf{Implementation Details.}
In the experiments, our rumor detector uses a pre-trained BERT model, with \textit{bert-base-uncased} for the English datasets and \textit{chinese-bert-wwm-ext} for the Chinese datasets. For the \textbf{comment generator tuning}, we use two versions of the generative models: \textit{flan-t5-base} and \textit{llama-3-8b}. During the tuning phase, we empirically use $L = 10$ experts for generation, where each expert employs the LoRA structure \citep{hu2022lora}. We also use the \textit{spaCy} toolkit for entity extraction from tweets and a pre-trained \textit{gpt2} model for generating entity descriptions. 
We train the generator with the AdamW optimizer, which has an adaptive learning rate, a batch size of 8, and the model is trained for 5 epochs by default. In the \textbf{detector training} phase, we use the Adam optimizer with a learning rate of $7 \times 10^{-5}$, a batch size of 64, and apply an early stopping strategy that means the model training stops if no improvement in the Macro F1 score is observed after 10 consecutive epochs. We repeat the training of each model five times using five different seeds $\{1, 2, 3, 4, 5\}$ and report their average results in subsequent experiments.
For the model’s hyper-parameters, we experimentally fix $\alpha$ and $\beta$ to 1, choose the threshold $\tau$ in the controversy fusion to be the median value in the $\xi$ sequence, and set $\epsilon = 0.5$.

\begin{table}[t]
\centering
\renewcommand\arraystretch{0.9}
  \caption{Style similarity and diversity of generated comments compared to three baseline methods across four RD datasets.}
  \label{diversity}
  \small
  \setlength{\tabcolsep}{5pt}{
  \begin{tabular}{m{1.20cm}<{\centering}m{0.52cm}<{\centering}m{0.52cm}<{\centering}m{0.52cm}<{\centering}m{0.52cm}<{\centering}m{0.52cm}<{\centering}m{0.52cm}<{\centering}m{0.52cm}<{\centering}m{0.52cm}<{\centering}}
    \toprule
    \multirow{2}{*}{Method} & \multicolumn{2}{c}{\textit{Twitter15}} & \multicolumn{2}{c}{\textit{Twitter16}} & \multicolumn{2}{c}{\textit{Weibo16}} & \multicolumn{2}{c}{\textit{Weibo20}} \\
    \cmidrule(r){2-3} \cmidrule(r){4-5} \cmidrule(r){6-7} \cmidrule(r){8-9}
    & Sty.$\uparrow$ & Div.$\downarrow$ & Sty.$\uparrow$ & Div.$\downarrow$ & Sty.$\uparrow$ & Div.$\downarrow$ & Sty.$\uparrow$ & Div.$\downarrow$ \\
    \hline
    \rowcolor{lightgrayv} \textbf{\baby} & \textbf{.8078} & \textbf{.8171} & \textbf{.8041} & \textbf{.7914} & \textbf{.6277} & \textbf{.7815} & \textbf{.6173} & \textbf{.7882} \\
    DELL & .8002 & .8544 & .7919 & .8417 & .5870 & .8183 & .5703 & .8016 \\
    GenFEND & .8026 & .8457 & .7955 & .8510 & .5811 & .8237 & .5635 & .8044 \\
    Prompt & .7938 & .8936 & .7852 & .8892 & .4938 & .8851 & .5001 & .8645 \\
    \bottomrule
  \end{tabular} }
\end{table}

\subsection{Comparative Experiments}

To evaluate the performance of comments generated by our CGT module and our overall \baby framework, we compare them with SOTA RED baselines and comments generation methods.

\subsubsection{Compared with RED Baselines} \label{sec4.2.1}

We validate our method in two RED scenarios: training on full comments and testing in an early scenario ($M = 16$ and $M^\prime = 2$), and both the training and test comments are limited ($M = 2$ and $M^\prime = 2$). We generate comments to keep $M + K = M^\prime + K^\prime = 16$. The experimental results are reported in Table~\ref{resultearly} and Table~\ref{resultcontent}, respectively. Generally, the comments generated by our CGT consistently and significantly improve the performance of the baseline models, and the results from our proposed comment integration module also outperform other baseline models. For example, on the \textit{Twitter16} dataset, our generated comments improve CAS-FEND by an average of 4.26 across all metrics.
We compare the performance of our method under two different scenarios. Under the scenario that the number of training comments exceeds the number of test comments, our method yields a greater improvement, with an average increase of 2.59 across all datasets and models. We observe that in this scenario, the overall performance of our model is even lower than when both the training and test comments are limited. However, this is due to the imbalance between training and test comments, which results in lower performance for the baseline models. This finding demonstrates that our method can address this imbalance issue by generating high-quality comments.

\subsubsection{Compared with Generation Methods} \label{sec4.2.2}

We conduct two experiments to compare the performance of our comment generation method with cutting-edge generation methods, DELL \citep{wan2024dell} and GenFEND \citep{nan2024let}. First, we report the improvements in the detection performance of the baseline model with different comment generation methods in Table~\ref{resultgeneration}. We select CAS-FEND \citep{nan2024exploiting} and our proposed comment integration module for the experiments.
In the experiments, we find that the Llama-based CGT consistently achieves the best performance, demonstrating that our proposed training approach effectively generates diverse, knowledgeable, and human-like comments, significantly improving the performance of the baseline models. Additionally, the performance of the other three generation methods is quite similar, and our T5-based model slightly outperforms the other two approaches based on the larger Llama model. This also demonstrates both the efficiency and effectiveness of our process.

Second, we design two metrics to measure the style similarity between the generated comments and the original comments, as well as their diversity, respectively. The experimental results are shown in Table~\ref{diversity}. Formally, given original comments $\{\mathbf{c}_{ij}\}_{j=1}^M$ and generated comments $\{\mathbf{c}_{ij}^\gamma\}_{j=1}^K$, we use pre-trained BERT to obtain their hidden embeddings $\{\mathbf{h}_{ij}\}_{j=1}^M$ and $\{\mathbf{h}_{ij}^\gamma\}_{j=1}^K$, and then calculate two metrics as follows:
\begin{align}
\small
    \text{Sty.} = \frac{1}{NMK} \sum \nolimits _{i=1}^N \sum \nolimits _{j=1}^M \sum \nolimits _{k=1}^K \frac{\mathbf{h}_{ij} \cdot \mathbf{h}_{ik}^\gamma}{\|\mathbf{h}_{ij}\| \times \|\mathbf{h}_{ik}^\gamma\|}, \nonumber \\
    \text{Div.} = \frac{1}{NK^2} \sum \nolimits _{i=1}^N \sum \nolimits _{j=1}^K \sum \nolimits _{k=1}^K \frac{\mathbf{h}_{ij}^\gamma \cdot \mathbf{h}_{ik}^\gamma}{\|\mathbf{h}_{ij}^\gamma\| \times \|\mathbf{h}_{ik}^\gamma\|}, \nonumber
\end{align}
where, a higher Sty. value indicates that the style of the generated comment is closer to that of the original comment, and a lower Div. value suggests higher diversity in the generated comments. The experimental results in Table~\ref{diversity} show that our generated comments outperform other generation methods in both style alignment with the original comments and diversity, while the performance of the direct prompting on the Llama model is the worst. This demonstrates the effectiveness of our proposed multiple expert structure and the adversarial style alignment module.

\begin{figure}[t]
  \centering
  \includegraphics[scale=0.22]{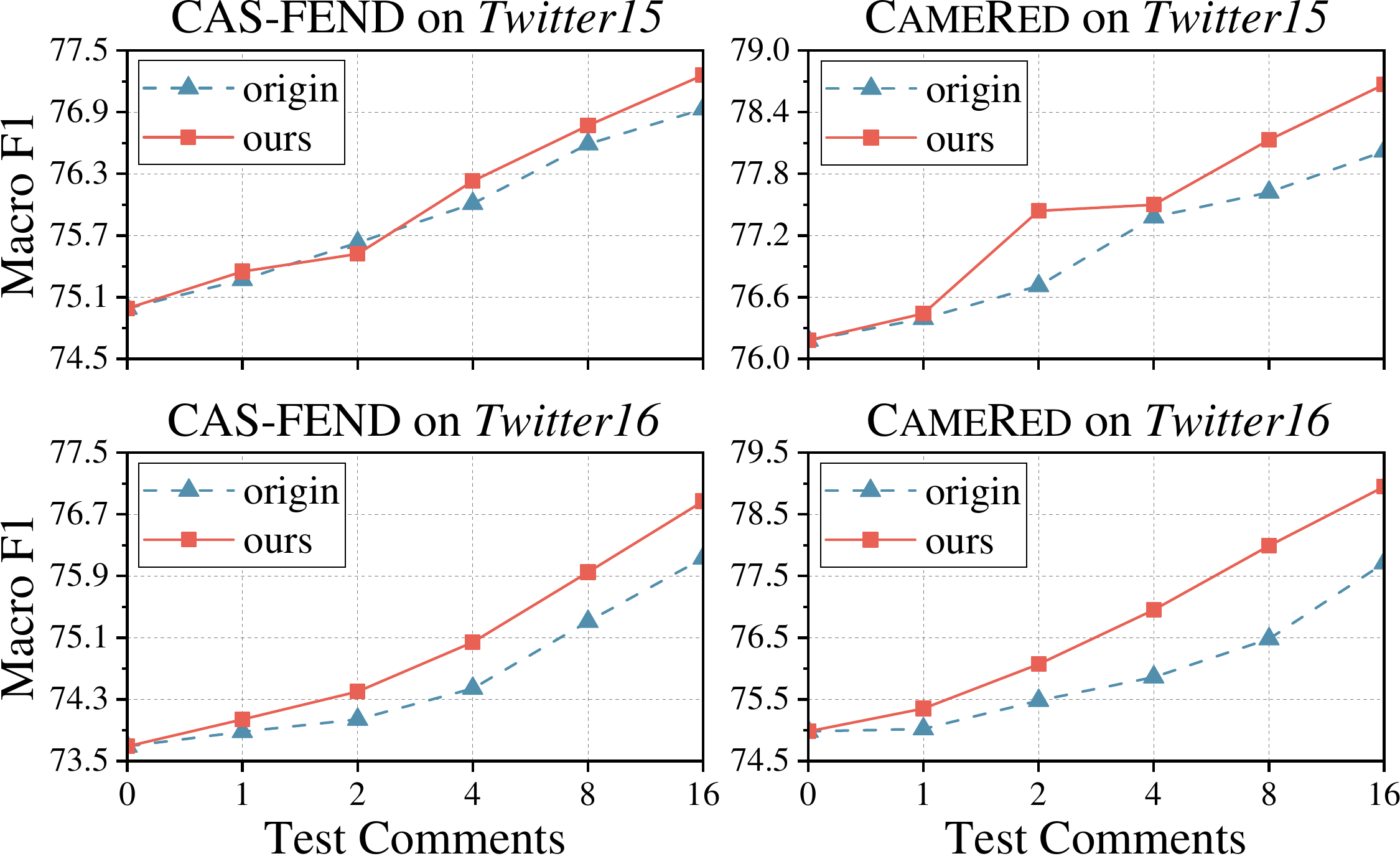}
  \caption{Comparison between original and our comments.}
  \label{compare_origin}
\end{figure}

\begin{table*}[t]
\centering
\renewcommand\arraystretch{0.98}
  \caption{Experimental results of \baby and SOTA generation methods.}
  \label{resultgeneration}
  \small
  \setlength{\tabcolsep}{5pt}{
  \begin{tabular}{m{3.3cm}m{0.67cm}<{\centering}m{0.67cm}<{\centering}m{0.67cm}<{\centering}m{0.67cm}<{\centering}m{0.67cm}<{\centering}m{0.67cm}<{\centering}m{0.67cm}<{\centering}m{0.67cm}<{\centering}m{0.67cm}<{\centering}m{0.67cm}<{\centering}m{0.67cm}<{\centering}m{0.67cm}<{\centering}m{0.94cm}<{\centering}}
    \toprule
    \multirow{2}{*}{\quad \quad Model} & \multicolumn{3}{c}{Dataset: \textit{Twitter15} \citep{ma2017detect}} & \multicolumn{3}{c}{Dataset: \textit{Twitter16} \citep{ma2016detecting}} & \multicolumn{3}{c}{Dataset: \textit{Weibo16} \citep{ma2017detect}} & \multicolumn{3}{c}{Dataset: \textit{Weibo20} \citep{ma2016detecting}} & \multirow{2}{*}{Avg.~$\boldsymbol{\Delta}$} \\
    
    \cmidrule(r){2-4} \cmidrule(r){5-7} \cmidrule(r){8-10} \cmidrule(r){11-13} 
    & Acc. & F1 & AUC & Acc. & F1 & AUC & Acc. & F1 & AUC & Acc. & F1 & AUC & \\

    \hline
    CAS-FEND \citep{nan2024exploiting} & 71.18 & 74.99 & 91.56 & 73.76 & 73.69 & 91.44 & 83.25 & 81.69 & 80.99 & 85.51 & 85.47 & 85.54 & - \\
    \rowcolor{lightgrayv} \quad + CGT \textit{w/} T5 \citep{chung2022scaling} & 77.92 & 78.09 & 92.79 & 78.04 & 77.43 & 92.51 & 83.75 & 82.47 & 82.08 & 86.86 & 86.86 & 86.86& \textbf{+2.22} \\
    \rowcolor{lightgrayv} \quad + CGT \textit{w/} Llama \citep{touvron2023llama} & 78.93 & 78.86 & 92.31 & 78.44 & 78.32 & 93.90 & 84.54 & 83.33 & 82.93 & 87.92 & 87.92 & 87.93 & \textbf{+3.02} \\
    \quad + DELL \textit{w/} Llama \citep{wan2024dell} & 77.80 & 77.64 & 92.72 & 77.86 & 77.26 & 91.20 & 83.85 & 82.48 & 81.92 & 86.84 & 86.83 & 86.86 & \textbf{+2.02} \\
    \quad + GenFEND \textit{w/} Llama \citep{nan2024let} & 77.92 & 77.88 & 92.76 & 77.86 & 77.25 & 91.83 & 83.86 & 82.55 & 82.13 & 87.03 & 87.03 & 87.03 & \textbf{+2.17} \\
    
    \hline
    \baby \textit{w/o} CGT & 76.43 & 76.18 & 91.92 & 75.06 & 74.98 & 92.92 & 83.72 & 82.66 & 82.60 & 86.56 & 86.55 & 86.56 & - \\
    \rowcolor{lightgrayv} \quad + CGT \textit{w/} T5 \citep{chung2022scaling} & 79.02 & 79.29 & 93.02 & 78.75 & 78.30 & 93.02 & 85.07 & 84.12 & 84.13 & 87.85 & 87.84 & 87.85 & \textbf{+1.84} \\
    \rowcolor{lightgrayv} \quad + CGT \textit{w/} Llama \citep{touvron2023llama} & 80.36 & 80.11 & 93.54 & 79.55 & 79.53 & 93.69 & 86.21 & 85.32 & 85.29 & 88.63 & 88.63 & 88.65 & \textbf{+2.78} \\
    \quad + DELL \textit{w/} Llama \citep{wan2024dell} & 78.90 & 79.07 & 92.45 & 78.68 & 78.45 & 92.86 & 85.01 & 84.00 & 83.87 & 87.65 & 87.65 & 87.65 & \textbf{+1.68} \\
    \quad + GenFEND \textit{w/} Llama \citep{nan2024let} & 79.22 & 79.30 & 92.62 & 78.83 & 78.46 & 92.51 & 85.09 & 84.10 & 84.00 & 87.80 & 87.80 & 87.80 & \textbf{+1.78} \\
    
    \bottomrule
  \end{tabular} }
\end{table*}

\begin{table}[t]
\centering
\renewcommand\arraystretch{0.98}
  \caption{Ablative results of five ablative versions.}
  \label{ablation}
  \small
  \setlength{\tabcolsep}{5pt}{
  \begin{tabular}{m{1.40cm}m{0.59cm}<{\centering}m{0.59cm}<{\centering}m{0.59cm}<{\centering}m{0.59cm}<{\centering}m{0.59cm}<{\centering}m{0.59cm}<{\centering}m{0.65cm}<{\centering}}
    \toprule
    \multirow{2}{*}{\quad Method} & \multicolumn{3}{c}{\textit{Twitter15}} & \multicolumn{3}{c}{\textit{Twitter16}} & \multirow{2}{*}{Avg. $\downarrow$} \\
    \cmidrule(r){2-4} \cmidrule(r){5-7}
    & Acc. & F1 & AUC & Acc. & F1 & AUC & \\
    \hline
    \rowcolor{lightgrayv} \textbf{\baby} & \textbf{80.36} & \textbf{80.11} & \textbf{93.54} & \textbf{79.55} & \textbf{79.53} & \textbf{93.69} & - \\
    \quad \textit{w/o} CGT & 76.43 & 76.18 & 91.92 & 75.06 & 74.96 & 92.92 & \textbf{-3.21} \\
    \quad \textit{w/o} $\mathcal{L}_{SA}$ & 79.29 & 78.91 & 92.94 & 78.85 & 78.78 & 92.97 & \textbf{-0.84} \\
    \quad \textit{w/o} $\mathcal{D}_\kappa$ & 79.43 & 78.63 & 93.01 & 78.92 & 78.93 & 92.85 & \textbf{-0.84} \\
    \quad \textit{w/o} HCR & 78.95 & 78.67 & 92.57 & 78.85 & 78.78 & 92.97 & \textbf{-1.00} \\
    \quad \textit{w/o} MCF & 79.29 & 78.91 & 93.24 & 78.22 & 78.40 & 93.77 & \textbf{-0.83} \\
    \hline
    \specialrule{0em}{0.5pt}{0.5pt}
    \hline
    \multirow{2}{*}{Method} & \multicolumn{3}{c}{\textit{Weibo16}} & \multicolumn{3}{c}{\textit{Weibo20}} & \multirow{2}{*}{Avg. $\downarrow$} \\
    \cmidrule(r){2-4} \cmidrule(r){5-7}
    & Acc. & F1 & AUC & Acc. & F1 & AUC & \\
    \hline
    \rowcolor{lightgrayv} \textbf{\baby} & \textbf{86.21} & \textbf{85.32} & \textbf{85.29} & \textbf{88.63} & \textbf{88.63} & \textbf{88.65} & - \\
    \quad \textit{w/o} CGT & 83.72 & 82.66 & 82.60 & 86.56 & 86.55 & 86.56 & \textbf{-2.34} \\
    \quad \textit{w/o} $\mathcal{L}_{SA}$ & 85.65 & 84.64 & 84.03 & 87.89 & 87.89 & 87.90 & \textbf{-0.79} \\
    \quad \textit{w/o} $\mathcal{D}_\kappa$ & 85.73 & 84.97 & 84.11 & 87.97 & 87.97 & 87.96 & \textbf{-0.67} \\
    \quad \textit{w/o} HCR & 85.20 & 84.11 & 83.98 & 87.91 & 87.90 & 87.93 & \textbf{-0.95} \\
    \quad \textit{w/o} MCF & 84.94 & 83.87 & 83.67 & 88.33 & 88.32 & 88.32 & \textbf{-0.88} \\
    \bottomrule
  \end{tabular} }
\end{table}

\subsection{Ablative Study}

To validate the effectiveness of each module in our \baby, we conduct an ablative study of five ablative versions by removing generated comments with CGT, adversarial style alignment loss $\mathcal{L}_{SA}$, synthesized knowledgeable dataset $\mathcal{D}_\kappa$, heuristic collaboration routing (HCR), and mutual controversy fusion (MCF) module, and report their ablative results in Table~\ref{ablation}.
Generally, the results of all ablation versions are lower than those of our complete method, demonstrating the effectiveness of each module in the approach. Furthermore, the performance of these ablation versions can be roughly ranked as \textit{w/o} CGT $<$ \textit{w/o} HCR $<$ \textit{w/o} $\mathcal{L}_{SA}$ $<$ \textit{w/o} $\mathcal{D}_\kappa$ $<$ \textit{w/o} MCF, highlighting their relative importance.
Specifically, after removing HCR, the generation of comments no longer relies on the collaboration between multiple experts, but instead, each expert independently generates comments. This significantly reduces the diversity of the generated comments, while the knowledge in the comments comes solely from one single expert, decreasing the amount of knowledge present and negatively affecting detection performance. Then, removing style alignment causes the language style of the comments to differ from the original comments, introducing noise that impacts detection performance. This will be further explored in Sec.~\ref{sec4.5.2}. Lastly, removing the synthesized knowledge dataset results in comments with less informative content, leading to more superficial responses that hinder improvements in detection results.

\begin{figure}[t]
  \centering
  \includegraphics[scale=0.285]{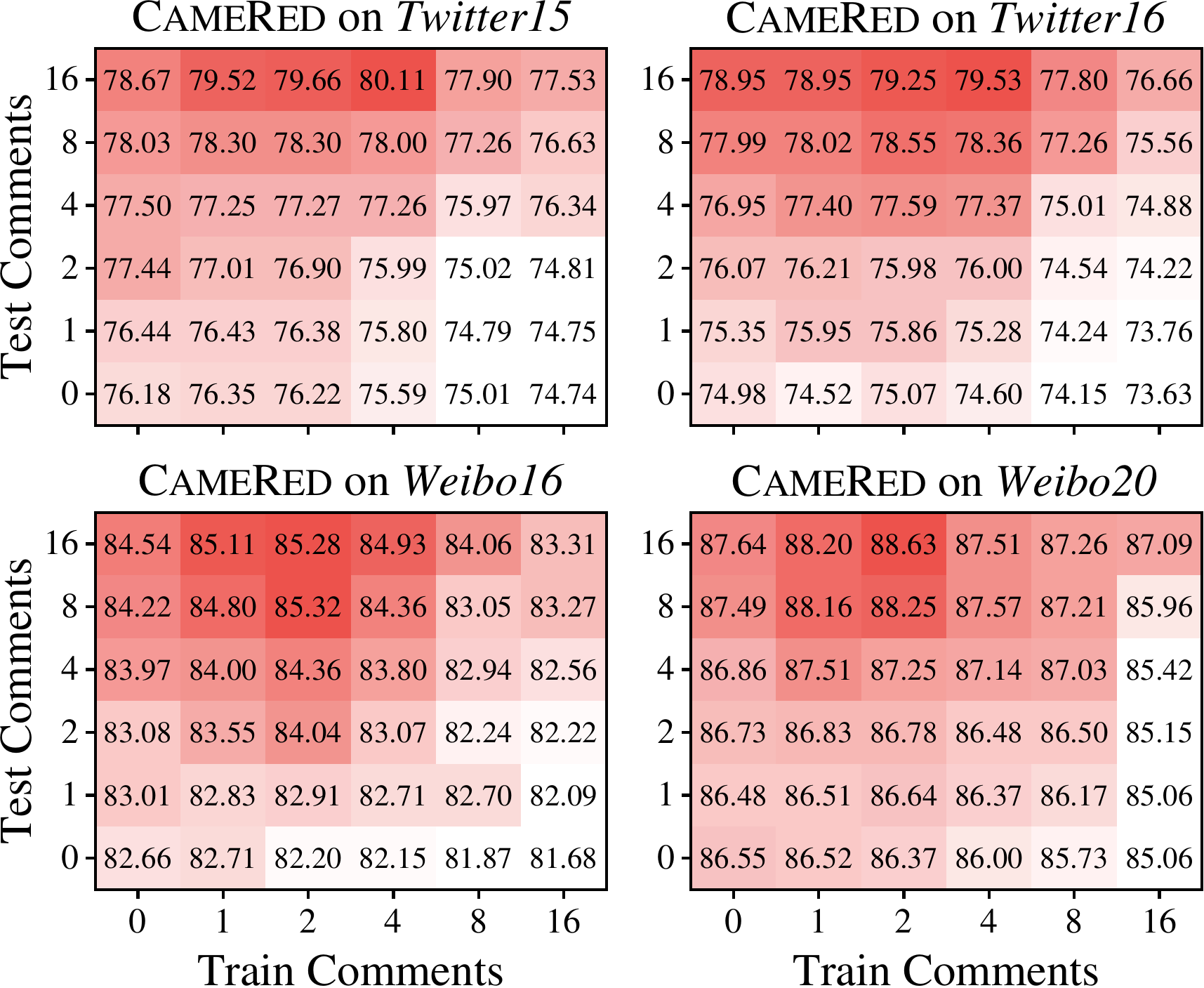}
  \caption{Sensitivity analysis of the number of train and test comments on four datasets.}
  \label{sensitivity}
\end{figure}

\begin{table*}[t]
\centering
\renewcommand\arraystretch{0.95}
  \caption{Three representative examples to illustrate the generated comments by DELL and our \baby.}
  \label{case}
  \footnotesize
  \begin{tabular}{m{5.5cm}|m{5.5cm}|m{5.5cm}}
    \bottomrule
    \rowcolor{lightgrayv} \textbf{Article}: One person dead, many taken to hospital after shootings, stabbing at Denver Coliseum, police say. & 
    \textbf{Article}: We hope you're as excited about Thursday as this kid was about LeBron's easy basket & 
    \textbf{Article}: Hackers from Anonymous say they're set to expose hundreds of KKK members \#OpKKK \\
    \rowcolor{lightbluev} \textbf{Veracity label}: \textit{Real} \quad \textbf{Source}: \textit{Twitter15} & 
    \textbf{Veracity label}: \textit{Real} \quad \textbf{Source}: \textit{Twitter16} &
    \textbf{Veracity label}: \textit{Fake} \quad \textbf{Source}: \textit{Twitter16}\\
    
    \hline
    \textbf{Original comments}: \ding{172} Gangs are responsible for way more crime and violence than terrorists &
    \textbf{Original comments}: \ding{172} Love seeing kids cheer for their teams with passion! &
    \textbf{Original comments}: \ding{172} Need to be absolutely sure the names they release are truly, incontrovertibly KKK \\
    \ding{173} Please people. We need to fix these gun control laws. Too many killings! & 
    \ding{173} The poor kid keeps looking back to see if the people he is w even notice him, and they don't! & 
    \ding{173} Seriously? What about the murders, lynchings tortures of people since Reconstruction? \\
    \ding{174} Truth is our National safety and economic strength is at risk. & 
    \ding{174} um...Correct me if I'm wrong but isn't JR Smith number 5? 5 is nowhere close to 23 lol &
    \ding{174} Good news. Hopefully they get all these people. Wonder if this one will be authentic. \\
    \hline
    \textbf{Comments generated by DELL} \citep{wan2024dell}: \ding{172} Breaking news! Tragic incident at the Denver Coliseum where one person has lost their life & 
    \textbf{Comments generated by DELL} \citep{wan2024dell}: \ding{172} This little dude's reaction to LeBron James' easy layup is everything! & 
    \textbf{Comments generated by DELL} \citep{wan2024dell}: \ding{172} Hacktivists from Anonymous claim they've obtained a massive database \\
    \ding{173} This is an absolute pity, but what did the cops not realize in the case of that shooting? & 
    \ding{173} This kids reaction to LeBron's layup is everything! Can we all just be this enthusiasm & 
    \ding{173} Hackers from Anonymous claim they've got access to personal information of over 50 \\
    \ding{174} The police takes a handful of calls and are rethinking a strategy to bring the victims to justice for the injuries & 
    \ding{174} This kids excitement is contagious! I love how much they care about sports and their favorite players. Great & 
    \ding{174} It's good he's got this thing but what are you gonna do next? This is all I ever know \\
    \hline
    \textbf{Comments generated by \baby}: \ding{172} We need a better way of handling these things in general and especially with children. & 
    \textbf{Comments generated by \baby}: \ding{172} He will have a great time watching the game next weekend with all these youngsters and their friends in school &
    \textbf{Comments generated by \baby}: \ding{172} No one has done that yet? And I think it's a scam so far! But there is no way for us to know what \\
    \ding{173} The first time in years the public had a chance to see how much we have lost since then it was just too late for us all and there & 
    \ding{173} The best team in the league and he is still a rookie great job by him well done Lebron James keep up good work guys &
    \ding{173} I don't think that would be a good idea there is enough hate in the world without making it easier for others to spread theirs too  \\
    \ding{174} What is going on with the country today why do these things keep happening gun laws need to change & 
    \ding{174} I know that it's a big day for all of us and the kids so let them enjoy their dream to meet some real heroes &
    \ding{174} I bet that will make the kkk very nervous lol thanks for posting it here on your blog post about hackers \\
    \bottomrule
  \end{tabular}
\end{table*}

\subsection{Sensitivity Analysis of Comments}

Our preliminary experiments demonstrate that an imbalance in the number of training and testing comments leads to a significant decline in the model's detection performance. To investigate whether our generated comments exhibit the same behavior, we provide a sensitivity analysis regarding the number of comments in Fig.~\ref{sensitivity}. We conduct experiments on our model \baby under the scenario that $M = 16$ and $M^\prime = 2$, varying the number of training and test comments to $\{0, 1, 2, 4, 8, 16\}$.
The experimental results consistently show that when the total number of comments (original and generated) is balanced between training and testing, i.e., $M + K \approx M^\prime + K^\prime$, the model performs optimally. In contrast, as the discrepancy between them increases, the model's performance exhibits a consistent decline, even falling below the performance of original baseline models.

\subsection{Evaluation of Generated Comments}

In this section, we evaluate generated comments by comparing them with original comments (Sec.~\ref{sec4.5.1}) and comments with corrupted styles (Sec.~\ref{sec4.5.2}), and present some representative cases (Sec.~\ref{sec4.5.3}).

\begin{figure}[t]
  \centering
  \includegraphics[scale=0.22]{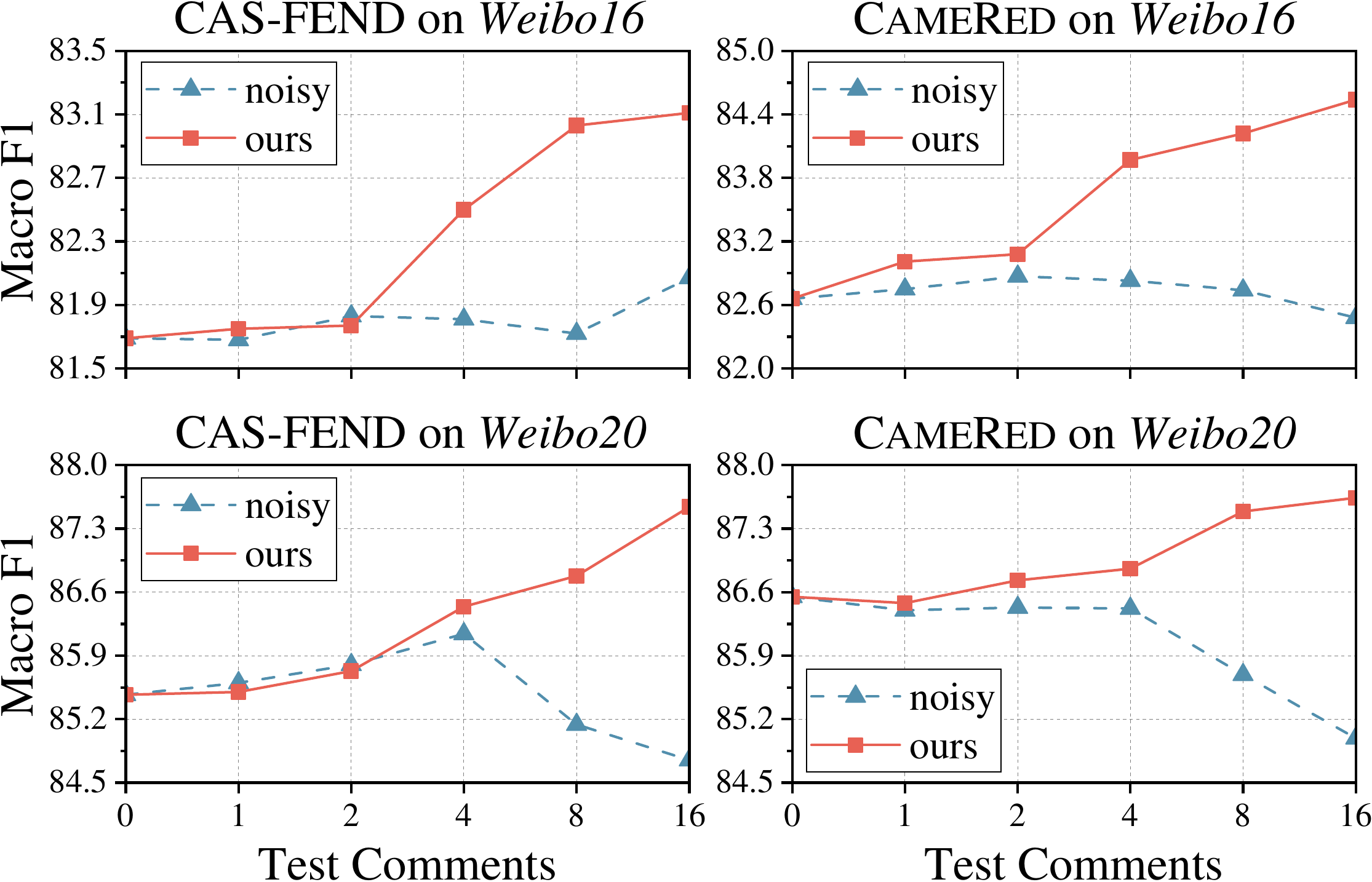}
  \caption{Comparison between noisy and our comment style.}
  \label{compare_style}
\end{figure}

\subsubsection{Compared with Original Comments} \label{sec4.5.1}

We first investigate the impact of our generated comments, compared to the original authentic user comments in the dataset, on detection performance. We conduct experiments in the scenario that $M = 16$ and $M^\prime = 2$, varying the number of test comments from $\{0, 1, 2, 4, 8, 16\}$. We choose CAS-FEND and \baby for evaluation across the \textit{Twitter15} and \textit{Twitter16} datasets, and the results are shown in Fig.~\ref{compare_origin}.
In general, the model's performance consistently improves as the number of test comments increases, as a higher number of test comments brings the training and test comments closer to being balanced. Additionally, our generated comments outperform the original comments in most settings. These findings demonstrate that our method can generate high-quality comments, even surpassing original comments, which is attributed to our ability to inject more knowledge into the comments and maintain their diversity.

\subsubsection{Compared with Corrupted Styles} \label{sec4.5.2}

To validate the necessity of style alignment and the effectiveness of our proposed style alignment method, we compare our generated comments with versions that corrupt their styles. We select CAS-FEND and \baby as baseline models and report their experimental results on \textit{Weibo16} and \textit{Weibo20} in Fig.~\ref{compare_style}. Additionally, we use the same RED scenario as in Sec.~\ref{sec4.5.1}, varying the number of test comments from $\{0, 1, 2, 4, 8, 16\}$. Specifically, we corrupt the style of the original comments by prompting an LLM Llama, to change the language style while maintaining the original meaning and length, making the comments more formal or playful. 
The experimental results show that as the number of test comments increases, our generated comments consistently improve model performance. However, an increase in style-corrupted comments leads to performance degradation, especially when the number of comments is higher, where the decline becomes more significant. These results demonstrate that aligning the style of the comments during the training of the comment generator is necessary, and that our method effectively aligns the style of the original comments.

\subsubsection{Case Study} \label{sec4.5.3}

To directly demonstrate our generated comments compared to the original comments and those generated by other methods, \eg DELL \citep{wan2024dell}, we provide three representative examples from the test subsets of \textit{Twitter15} and \textit{Twitter16}.
Generally, our method generates comments that are more informative and diverse. For example, in the second case, DELL consistently focuses on "\textit{layup}," which limits its diversity, whereas our method generates comments from more varied perspectives. It also knows that \textit{LeBron James} is a rookie in the league, a detail not mentioned in the original article, indicating that our model is better at capturing both the inherent knowledge within the model and the potential knowledge in the synthesized training data.

\section{Conclusion and Limitations}

In this paper, we concentrate on detecting rumors in their early stage. We empirically observe that the detection models perform better when both training and test comments are extensive. Upon the observations, we propose a new framework \baby to generate human-like comments by tuning a comment generator. We specify the tuning by simulating the collaboration and controversy among human experts. Specifically, we inject an MoE structure into a pre-trained language model, and design a heuristic collaboration routing strategy to generate diverse comments and synthesize a new dataset to make comments knowledgeable. To control the style of comments, we present an adversarial style alignment method. Extensive experiments demonstrate the performance of \baby.

\section*{Acknowledgement}
We acknowledge support for this project from the National Key R\&D Program of China (No.2021ZD0112501, No.2021ZD0112502), the National Natural Science Foundation of China (No.62276113), and China Postdoctoral Science Foundation (No.2022M721321).

\clearpage
\bibliographystyle{ACM-Reference-Format}
\balance
\bibliography{reference}


\end{document}